%% file: main_arxiv.tex
\documentclass[]{fairmeta} %

\usepackage{graphicx} %
\usepackage{comment}
\usepackage{palatino}
\usepackage{pifont}
\usepackage{graphicx}
\usepackage{amsmath}
\usepackage{amsfonts}
\usepackage{enumitem}
\usepackage{booktabs}
\usepackage[dvipsnames]{xcolor}
\usepackage{colortbl}
\usepackage{xspace}
\usepackage[font=small]{caption}
\title{\scalebox{1}{Short window attention } \newline \scalebox{1}{enables long-term memorization}}

\author{Loïc Cabannes$^{1,2,3}$}
\author{Maximilian Beck$^{1,4}$}
\author{Gergely Szilvasy$^1$}
\author{Matthijs Douze$^1$}
\author{Maria Lomeli$^1$}
\author{\mbox{Jade Copet}$^1$}
\author{Pierre-Emmanuel Mazar\'e$^1$}
\author{Gabriel Synnaeve$^1$}
\author{Herv\'e J\'egou$^1$}

\affiliation{$^1$Meta FAIR}
\affiliation{$^2$ENS Paris Saclay}
\affiliation{$^3$Paris Cité University}
\affiliation{$^4$Johannes Kepler University Linz}
\abstract{

\input{abstract}
}

\usepackage{xcolor}
\definecolor{darkgreen}{rgb}{0.0, 0.8, 0.13} %

\makeatletter
\DeclareRobustCommand\onedot{\futurelet\@let@token\@onedot}
\def\@onedot{\ifx\@let@token.\else.\null\fi\xspace}

\makeatother

\begin{document}
\maketitle
\bigskip \bigskip

\input{splash_fig}

\input{text_body}

\bibliography{refs}
\bibliographystyle{iclr2026_conference}

\newpage

\begin{center}
{\LARGE \textsc{Supplementary material}}
\end{center}

\appendix

\input{appendix}

\end{document}

%% file: abstract.tex
Recent works show that hybrid architectures combining local sliding window attention layers and global attention layers outperform either of these architectures taken separately. However, the impact of the window length and the interplay between local layers and global layers remain under-studied. In this work, we first analyze the interaction between short and long term memory by considering SWAX: a hybrid architecture consisting of sliding-window attention and xLSTM linear RNN layers.

A counter-intuitive finding is that larger sliding windows hurts the long-context performance. In fact, short window attention encourages the model to better train the long-term memory of the xLSTM as it cannot rely on the local softmax attention mechanism for long context-retrieval. 
We also validate our findings on local-global architectures alternating short window and full attention layers: the short layers should be small in order not to hinder the usefulness of the long layers. 

However, employing too small sliding windows is detrimental even for short-context tasks, which could be solved with information from moderately larger sliding windows otherwise. Therefore, we train hybrid architectures by stochastically changing the sliding window size, forcing the model to leverage both the short term window and the long-term memory. Training with stochastic window sizes significantly outperforms regular window attention both on short and long-context problems.

%% file: splash_fig.tex
\begin{minipage}{0.43\linewidth}
\includegraphics[width=\linewidth]{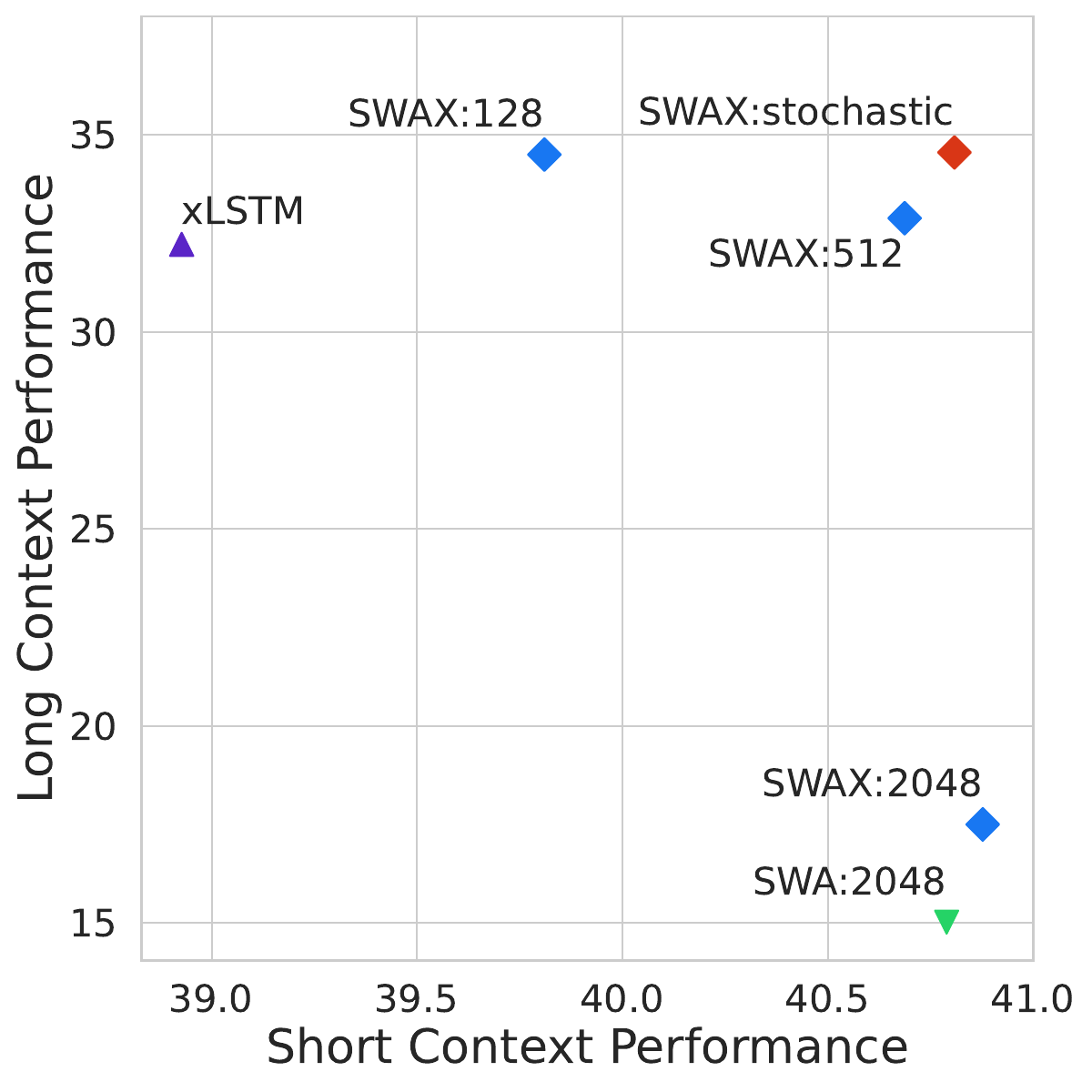}
\captionof{figure}{Short (average score across benchmarks) vs long context performance for 1.4B xLSTM, SWA (sliding window attention) and SWAX with different sliding window sizes.
\label{fig:splash1}
}
\end{minipage}
\hfill
\begin{minipage}{0.43\linewidth}
\includegraphics[width=\linewidth]{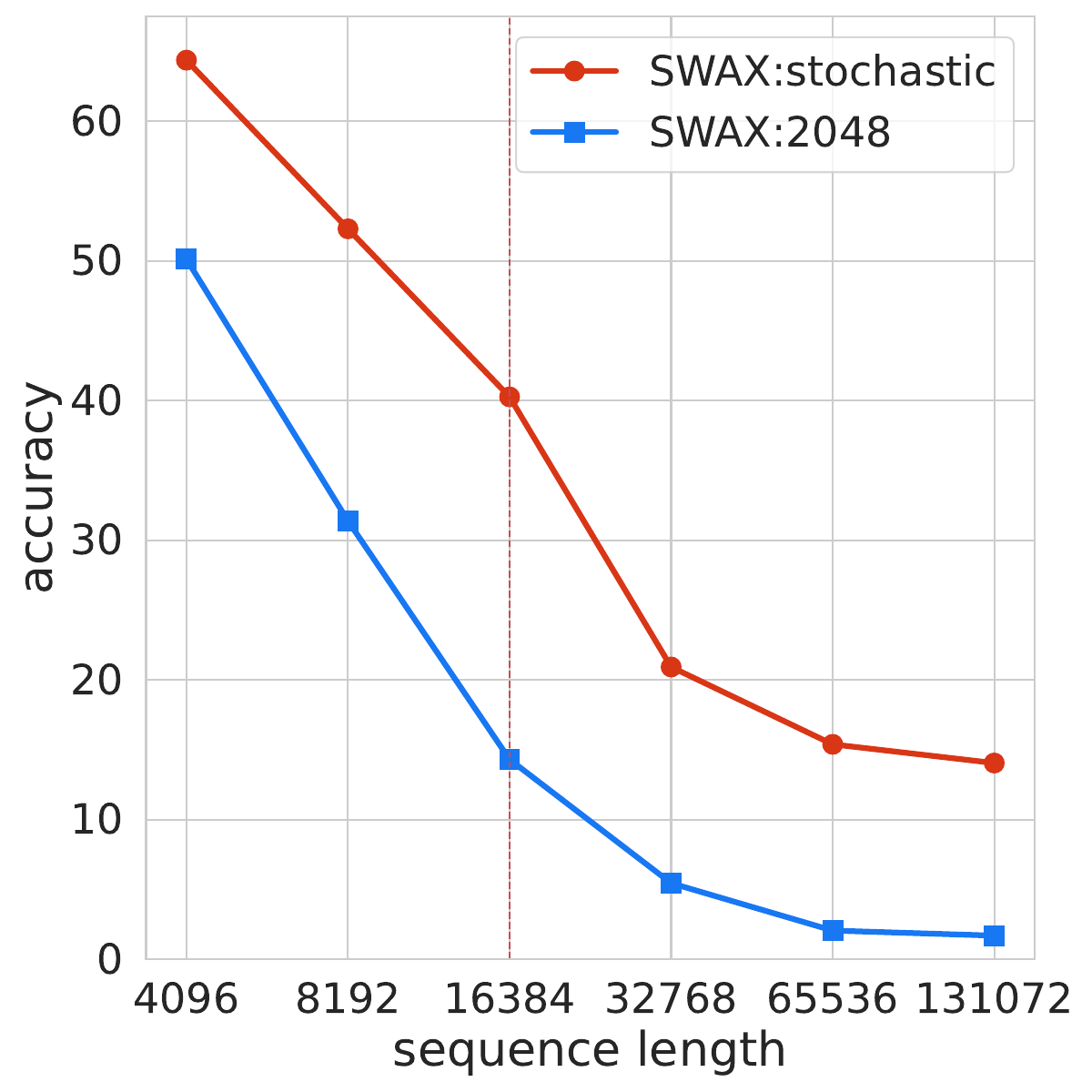}
\captionof{figure}{RULER Needle-In-A-Haystack accuracy of a 1.4B SWAX model with a fixed sliding window size of 2048 vs our method using a stochastic window size of 128/2048.
\label{fig:splash2}
}
\end{minipage}

%% file: text_body.tex
\section{Introduction}
Memory is a core concept in Neural Network Architectures \citep{zhong2025understandingtransformerperspectiveassociative}.
Modern LLMs based on softmax attention have a working memory in the form of the key-value (KV) Cache and yield state-of-the-art long-context performance. This working memory expands indefinitely as the sequence length grows, incurring a linear growth in both compute and memory to generate each new token.
With such an unbounded compute cost, current models become prohibitively expensive for in-context learning on long sequences such as codebases and long reasoning traces.

On the other hand, recurrent neural networks (RNNs) like State Space Models (SSMs) \citep{gu2024mambalineartimesequencemodeling} or variants of linear attention (LA) \citep{katharopoulos2020transformersrnnsfastautoregressive} maintain and iteratively update a hidden state.
Through input-dependent update rules, RNNs manage to decide whether to keep previous or add new information \citep{LSTM,chung2014empiricalevaluationgatedrecurrent}.
In this way, the compute and memory cost is constant and independent of the sequence length.
This allows models to learn at test time from large sequence lengths and reason without a specific token limit.
Recently, these linear RNNs have been generalized in the context of online learning \citep{liu2024longhornstatespacemodels,sun2025learninglearntesttime}.
However, the recall ability of linear RNNs remains inferior to that of Transformers \citep{fu2023hungryhungryhipposlanguage}.
This shortfall has hindered their adoption in favor of global attention-based architectures, which remain the current state-of-the-art architecture for language and code models \citep{jain2024livecodebench}.

Recent works like those of \citet{de2024griffinmixinggatedlinear}, \citet{ren2025sambasimplehybridstate}, \citet{dong2024hymbahybridheadarchitecturesmall} and \citet{arora2025simplelinearattentionlanguage} have aimed at combining the advantages of softmax attention and Linear Attention into hybrid architectures~\citep{wang2025systematicanalysishybridlinear}.
Following this line of research, in this paper, we study hybrid architectures, which combine linear RNNs and \emph{sliding window} attention -- both components with fixed maximum state size and thus fixed compute cost per token.

In this context, we make the following contributions:
\begin{enumerate}
    \item We study the impact of the sliding window length on a wide range of tasks, encompassing validation perplexity, short-context reasoning, common sense benchmarks, and long-context modeling tasks;
    \item We show that, contrary to previous belief, for hybrid architectures that interleave sliding window attention and linear RNNs, \textit{longer} sliding windows actually \textit{hurt} performance in long-context retrieval tasks compared to using \textit{shorter} windows;
    \item We present a training strategy based on a stochastic window size that achieve the best of both worlds: we attain the long-context performance enabled by short windows. At the same time the performance is on par for short-context and reasoning tasks associated with longer windows.
\end{enumerate}

\section{Background}

The attention mechanism handles sequences of key vector $\mathbf{k}_t\in \mathbb{R}^{d_{qk}}$ and value vectors $\mathbf{v}_t \in \mathbb{R}^{d_{v}}$.
A fundamental perspective proposed by \cite{katharopoulos2020transformersrnnsfastautoregressive} is that all forms of attention update a matrix memory by adding to it the outer product of
 key vector $\mathbf{k}_t\in \mathbb{R}^{d_{qk}}$ and  value vectors $\mathbf{v}_t \in \mathbb{R}^{d_{v}}$.
This holds, provided that we can apply a vector mapping to each of these vectors.
Then, to read from the memory, a query $\mathbf{q}_t\in \mathbb{R}^{d_{qk}}$ is compared to the previous keys using a similarity metric, usually the inner product $\langle \mathbf{q}, \mathbf{k} \rangle$.
In order to improve the accuracy of subsequent retrieval operations, a pre-processing feature mapping $\phi$ is applied to the keys and queries.
Defining the memory tensor as
\begin{equation}
\mathbf{H}_t \;=\; \sum_{t=1}^S \phi(\mathbf{k}_t)\, \mathbf{v}_t^\top \;\;\in\;\mathbb{R}^{d_{qk}\times d_v},\qquad
\mathbf{z}_t \;=\; \sum_{t=1}^S \phi(\mathbf{k}_t) \;\;\in\;\mathbb{R}^{d_{qk}},
\label{eq:attention-write}
\end{equation}
a normalized read is performed as
\begin{equation}
\mathbf{y}_t \;=\; \frac{\phi(\mathbf{q}_t)^\top \mathbf{H}_t}{\phi(\mathbf{q}_t)^\top \mathbf{z}_t}
\;=\; \frac{\sum_{i\le t} \langle \phi(\mathbf{q}_t),\phi(\mathbf{k}_i)\rangle\, \mathbf{v}_i}{\sum_{i\le t} \langle \phi(\mathbf{q}_t),\phi(\mathbf{k}_i)\rangle}.
\label{eq:attention-read}
\end{equation}

\paragraph{Linear attention.}
Equation~(\ref{eq:attention-write}) shows that if the kernel $\phi$ is a finite-dimensional mapping, then the feature-mapped keys as well as the memory tensor are also finite-dimensional and can be materialized and cached for future retrievals $(\mathbf{H}_t,\mathbf{z}_t)$ in \emph{constant} memory:
\begin{equation}
    \mathbf{H}_t \leftarrow \mathbf{H}_{t-1} + \phi(\mathbf{k}_t)\, \mathbf{v}_t^\top,\qquad
    \mathbf{z}_t \leftarrow \mathbf{z}_{t-1} + \phi(\mathbf{k}_t).
\label{la-write}
\end{equation}

All the keys and values are thus stored in constant memory.
The per-token read cost is  $O(d_{qk}\times d_v)$. Importantly, it does not depend on the sequence length $S$.

\paragraph{Softmax attention (SA).}
\cite{katharopoulos2020transformersrnnsfastautoregressive} show that softmax attention can be seen as performing the attention operation defined in Equations~\ref{eq:attention-write} and \ref{eq:attention-read}, i.e., as writing outer products between keys and values in a memory. In such a case, the keys and queries undergo an infinite-dimensional feature mapping induced by the exponential kernel in softmax attention.
Compared to linear attention, an infinite-dimensional exponential feature map reduces the interference between the stored keys and yields an improved retrieval accuracy.
Another consequence is that the memory $\mathbf{H}_t$ cannot be materialized and cached.
Instead, one needs to maintain \emph{all} the previous keys and queries in memory in order to compute the exponential of the dot products of the keys and queries, also referred to as the ``KV Cache''.
A well known issue inherent to the self-attention mechanism, is that the KV Cache size (and per token computation) increases linearly with the sequence length.

\paragraph{Gated linear attention.}
Another way to limit interference between keys in the sequence is to learn when to ``forget'' information and remove it from the memory.
This is the idea behind Gated Linear Attention \citep{yang2024gatedlinearattentiontransformers}, which improves stability and long-context performance through selective retention/forgetting of the information.
Let $\boldsymbol{\alpha}_t, \boldsymbol{\beta}_t, \boldsymbol{\lambda}_t \in \mathbb{R}^{d_{qk}}$ be write-, read-, and decay-gates or, equivalently, broadcastable vectors.
Gating is often implemented by learned affine maps and element-wise sigmoids.
The update and reading rule are as follows:
\begin{align}
\mathbf{H}_t &= \operatorname{diag}(\boldsymbol{\lambda}_t)\, \mathbf{H}_{t-1} + \operatorname{diag}(\boldsymbol{\alpha}_t)\, \boldsymbol{\phi}(\mathbf{k}_t)\, \mathbf{v}_t^\top, \label{eq:gla-write} \\
\mathbf{y}_t &= \big(\operatorname{diag}(\boldsymbol{\beta}_t)\, \boldsymbol{\phi}(\mathbf{q}_t)\big)^\top \mathbf{H}_t.
\end{align}
Gated Linear Attention as well as other modern RNNs remove the normalizing constant and, instead, rely on normalizing layers such as LayerNorm \citep{ba2016layernormalization} and RMSNorm \citep{zhang2019rootmeansquarelayer} in the network to stabilize training \citep{beck:25tfla}.

\paragraph{Sliding window attention (SWA).}
Softmax attention maintains all past $(k_i,v_i)$ pairs, producing linear growth in memory and compute with $t$ due to the KV cache. Variants with a sliding window of size $w$  restrict the attention process to only the previous $w$ tokens,
changing the memory and time complexity per-token from $O(S)$ to $O(w)$ complexity~\citep{beltagy2020longformerlongdocumenttransformer}.
This theoretically allows SWA architectures to handle arbitrarily large input sequences. In practice the receptive field of the model is limited to $O(lw)$ where $l$ is the number of SWA layers in the model. Moreover, it is unlikely that the theoretical receptive field is fully utilized in practice \citep{xiao2025sliding}.

\paragraph{Hybrids between local attention and global softmax attention.}
 Through multi-turn interactions \citep{gehring2025rlefgroundingcodellms}, tool-use or long Chain-Of-Thought reasoning \citep{wei2023chainofthoughtpromptingelicitsreasoning,deepseekai2025deepseekr1incentivizingreasoningcapability}, the length which models have to process has grown from a few thousands of tokens to tens or hundreds of thousands of tokens.
 This motivates several recent works \citep{openai2025gptoss120bgptoss20bmodel,dong2024hymbahybridheadarchitecturesmall,nvidia2025nemotronhfamilyaccurateefficient,ren2025sambasimplehybridstate} to consider new architectures whose computational cost grow less rapidly relative to sequence length than that of global softmax attention, while still performing well on long-context tasks.
One such type of architectures are hybrids, for which most layers have a fixed state size like sliding-window or Linear Attention Layers, and the rest of the layers are global softmax attention layers.
 However, because those architectures still keep some global attention layers to remain competitive on long-context tasks, they also keep the $O(S)$ scaling in state size and FLOPs per token.

\paragraph{Hybrids between linear attention and sliding window softmax attention.}
Another kind of hybridization involves only component with a fixed state size, as considered by \cite{de2024griffinmixinggatedlinear} and \cite{ren2025sambasimplehybridstate}, who hybridize linear attention variants with sliding window attention.
SWA paired with linear attention provides a natural split: the linear path maintains a compressed working memory with an unlimited receptive field; the windowed softmax path offers high-fidelity local reasoning.
Moreover, they demonstrate that, despite having \textit{fewer} LA layers which are the only ones with an unlimited receptive field, the long-context performance of such hybrids is actually \textit{higher} than that of a purely Linear Attention architecture.
In particular, \citep{de2024griffinmixinggatedlinear} investigated the impact of the size of the sliding window on validation perplexity. They found that longer windows yield better performance, making the choice of window size a purely a trade-off between performance and compute. However, they did not investigate the impact of the sliding window length on the \textit{long-context} performance of the models.
\section{Hybrid architecture design}

In this work, we focus on hybrid architectures that alternate sliding window attention and linear RNNs.
As a candidate for the linear RNN component, we choose the xLSTM~\citep{beck2024xlstmextendedlongshortterm}, as this architecture has been scaled to models having up to 7B parameters and has shown strong performance in a wide variety of tasks. Importantly, fast and efficient Triton kernels are available~\citep{beck2025xlstm7brecurrentllm, beck:25tfla}.
xLSTM introduced two novel memory cells: the sLSTM with a scalar memory and the mLSTM with matrix memory.
However, on language tasks the mLSTM cell shows superior performance over the sLSTM, which has been abandoned in the latest 7B xLSTM model.
We follow this choice and rely solely on the mLSTM cell in our hybrid architecture.
Subsequently, we use xLSTM and mLSTM interchangeably to refer to the same architecture.

\begin{figure}[t]
\begin{minipage}{0.45\linewidth}
    \raisebox{1.5em}{\includegraphics[width=1.0\linewidth]{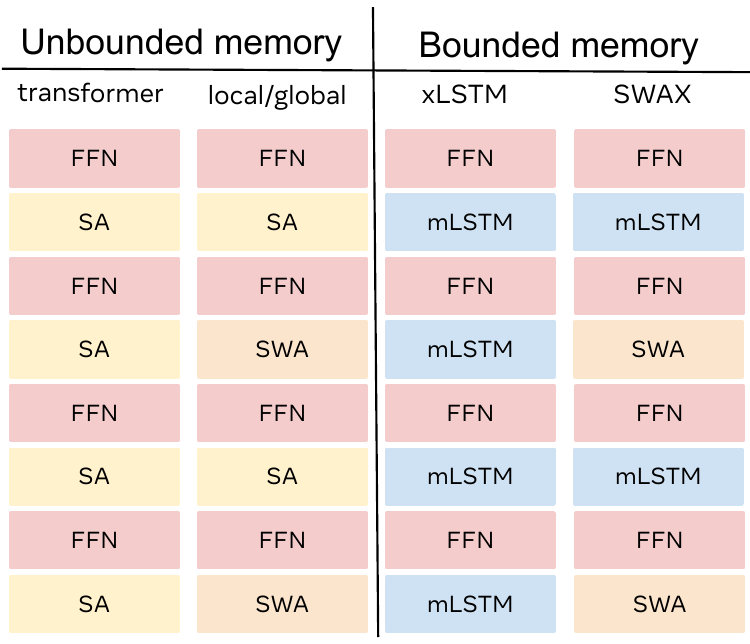}}
\end{minipage}
\hfill
\begin{minipage}{0.45\linewidth}
        \caption{We compare 4 different types of architectures, including 3 hybrid architectures:
        \medskip
        \newline (1) The transformer with vanilla self-attention (SA). Its  complexity is prohibitive for long contexts lengths.
        \medskip
        \newline (2) This is circumvented by replacing some SA layers by sliding window attention (SWA) layers \citep{gemmateam2025gemma3technicalreport,openai2025gptoss120bgptoss20bmodel}.
        \medskip
        \newline (3) xLSTM \citep{beck2024xlstmextendedlongshortterm} offers a memory with unbounded time horizon, albeit not as precise as SA for handling the recent context.
        \medskip
        \newline (4) SWAX is an hybrid architecture that includes both SWA layers and long-term memories layers, implemented with mLSTM memory cells.
    \label{fig:hybrid arch}}
\end{minipage}
\vspace{-1.5em}
\end{figure}

There exist many ways to hybridize these two components \citep{wang2025systematicanalysishybridlinear}.
In our case, we adopt the simple design of inter-layer hybridization, which alternates between SWA layers and layers of linear attention.
For the sake of simplicity, we adopt a 1:1 ratio meaning, that for every xLSTM layer there is one Sliding Window Attention layer.
Figure \ref{fig:hybrid arch} illustrates how the layers are interleaved in pure architectures and our SWA-xLSTM hybrid architecture.
Most hybrids use window sizes between 128 as in \cite{openai2025gptoss120bgptoss20bmodel} and 2048 as in \cite{de2024griffinmixinggatedlinear}. We evaluate at intermediate lengths with sliding attention windows of lengths 128, 256, 512, 1024 and 2048.
Finally, we evaluate a stochastic training procedure that aims at improving length-extrapolation.
This training procedure stochastically chooses for each new batch either a short or a long window.
In our experiments, we sample a window size of either 128 or 2048 with probability 0.5 for each length.
A similar strategy was proposed by \cite{zhang2025memory} in the context of the Memory Mosaic architecture.
However, in their case, the stochastic attention mask was applied to a long-term memory layer.
In constrast, we apply it to a Sliding Window Attention layer with the explicit goal of reducing over-reliance on the SWA layers for long-context recall.

\section{Experiments}
\subsection{Experimental setup}
Our experiments focus on language modeling, with an emphasis on understanding the compromise between short-context and long-context recall performance.
In particular, we investigate the impact of the SWA window size on long-context retrieval.
For this purpose, we mainly rely on the needle-in-a-haystack tasks of the RULER benchmark \citep{hsieh2024rulerwhatsrealcontext}.
A common practice for models using global attention is to pre-train them on shorter sequence lengths like 4k or 8k to reduce the cost of the attention operation, and then fine-tuned in a second training stage on longer sequences to improve their long-context ability \citep{peng2023yarnefficientcontextwindow}.
In our case, we mainly focus on fixed-memory, fixed-compute architectures. Therefore, longer training sequences do not increase the required compute to attain a total training tokens target. We choose to train our models on a 16k sequence length from the start.
Since we are interested in the capabilities of the model after standard pre-training, we do not perform any task-specific fine-tuning on long-context tasks.
Except stated otherwise, our experiments use a model with 1.4 billion parameters.
From our observations, it is at this size that models become able to perform recall on sequence lengths in the tens of thousands of tokens.
However, to validate our method of stochastic window size at larger scale, we also evaluate models at the 7B parameter scale.

The 1.4B models have 24 blocks and a model dimension of 2048 while the 7B models have 32 blocks and a model dimension of 4096.
In each block, the FFN is a gated MLP~\cite{liu2021payattentionmlps} with Silu activation~\citep{elfwing2017sigmoidweightedlinearunitsneural}.
For the SWA layers of the hybrids, we use Rotary Positional Embedding (RoPE) \citep{su2023roformerenhancedtransformerrotary} with a frequency $\theta$ of 10000, and 16 attention heads.
All models are trained on 150 billion tokens following a warmup-cosine learning rate schedule with a peak learning rate of $3\cdot10^{-4}$ and a minimum learning rate of $3\cdot10^{-6}$. The batch size is $10^6$ tokens.
Our training data mix consists mostly of web-data and code.
Since we are most interested in the performance of models on long sequences, and our code data has, on average, 10 times longer documents than our web data, we report the validation perplexity on the code data subset of our data mix.
For the MBPP \citep{austin2021programsynthesislargelanguage} and HumanEvalPLus \citep{evalplus} pass@10 results, we use a sampling temperature of 0.8.

\subsection{Hybrid \emph{vs.} pure architectures}

We start our investigation by reproducing the finding from \citep{de2024griffinmixinggatedlinear} whose hybridization of Local Attention and Linear Attention variants improve performance across the board on both short-term reasoning and long-term recall tasks.

\paragraph{Long-context performance.}
\begin{figure}[t]
    \centering
\includegraphics[width=0.5\linewidth]{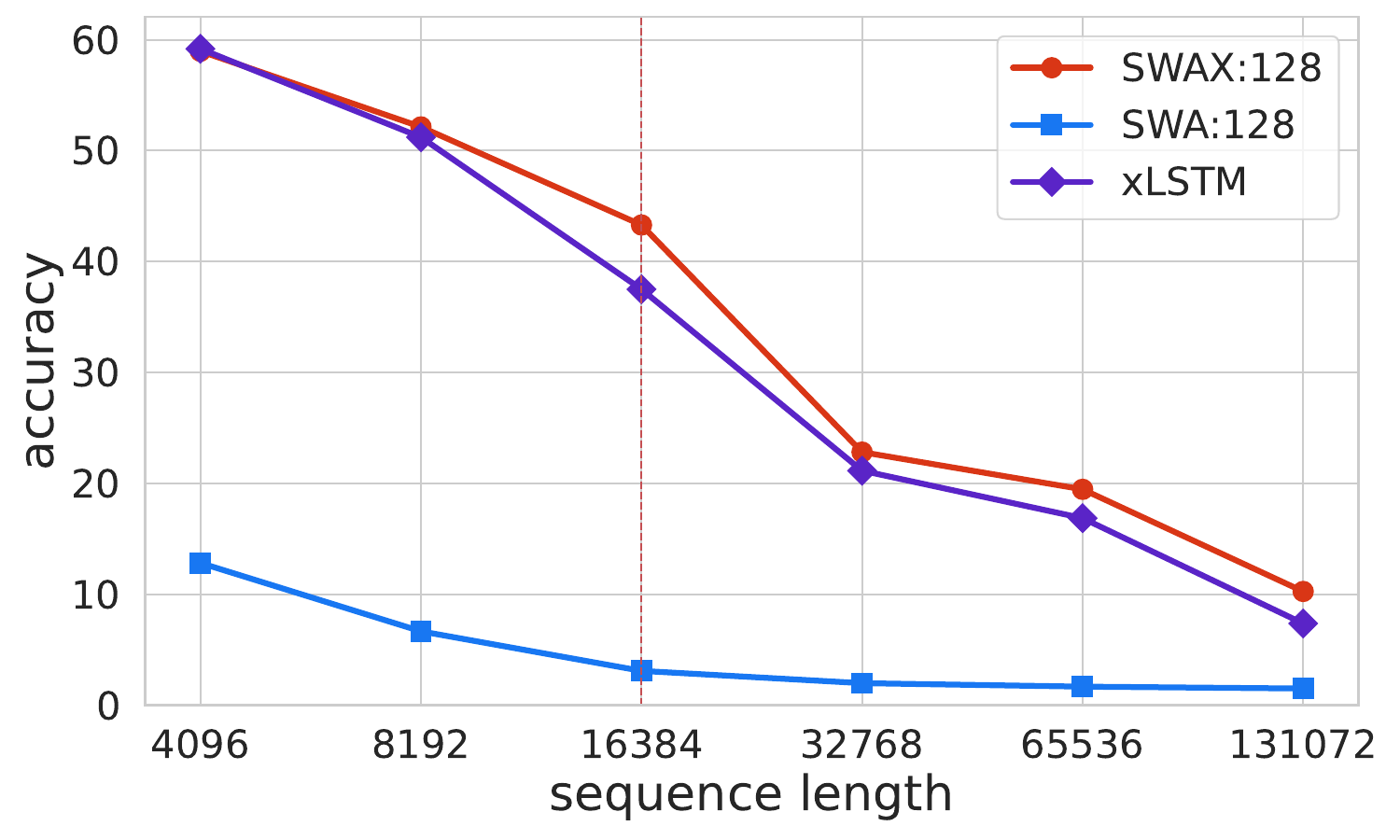}
    \caption{RULER needle-in-a-haystack average performance on varying sequence lengths for 1.4B models.
    For SWA and SWAX we indicate the sliding window size after the colon.}
    \label{fig:hybrids_base_RULER}
\end{figure}

Figure \ref{fig:hybrids_base_RULER} shows that the pure SWA architecture performs poorly on long-context recall.
This is expected because of its limited receptive field of 128 * 24 = 3072 tokens.
More importantly, it confirms the counterintuitive finding from \citep{de2024griffinmixinggatedlinear} that hybrids, despite having fewer global receptive field layers, outperform the pure variants in long-context recall.
Intuitively, this is explained by the fact that, although the SWA layers have a limited receptive field, the softmax feature mapping allows them to better model local dependencies than an equivalent number of linear attention layers.
Since most of the information necessary to predict the next token comes from local dependencies \citep{ruiz2025understandingimprovinglengthgeneralization}, a fully linear attention model dedicates most of its layers to modeling local dependencies and few layers to model long-term dependencies.

On the other hand, in SWA-LA hybrids, local dependencies are rather routed to the softmax local attention layers, which are more precise because of their direct access to recent history. As a consequence the linear attention layers specialize in modeling long-term dependencies, which the SWA layers cannot model due to the limited window size.
This highlights the impact that the window size can have on how much supervision the linear attention layers receive.

\paragraph{Short-context performance}
\begin{table}[t]
    \centering
\scalebox{0.85}{
    \begin{tabular}{lc@{\ \ }c@{\quad\quad}c|ccccc}
    \toprule
    Model                               & Transformer         & xLSTM & SWA & \multicolumn{5}{c}{SWAX} \\
    \midrule
    window length                       & n/a             & n/a   & 128            & 128               & 256            & 512            & 1024           & 2048 \\
    FLOPs/token ($\times 10^9$)        & 6.174           & 2.978 & 3.029          & 3.004             & 3.016          & 3.041          & 3.092          & 3.192 \\
    \midrule
    val\_PPL {\small$\downarrow$}         & \textbf{2.431}  & 2.602 & 3.036          & 2.551             & 2.540          & 2.546          & 2.538          & \textbf{2.523} \\
    \midrule
    HEplus/pass@10 {\small$\uparrow$}   & 14.63           & 12.80 & 13.41          & 12.20             & 12.80          & 14.02          & 14.63          & \textbf{15.24} \\
    ARC-c {\small$\uparrow$}            & 30.90           & 28.93 & 31.25          & 29.27             & 30.13          & \textbf{31.76} & 30.30          & 29.79          \\
    ARC-e {\small$\uparrow$}            & 66.43           & 65.79 & 65.58          & 65.75             & \textbf{67.82} & \textbf{67.82} & 66.89          & 67.15          \\
    Hellaswag {\small$\uparrow$}        & 46.18           & 44.68 & 44.45          & 45.37             & 45.37          & 45.47          & 45.47          & \textbf{45.91} \\
    MBPP/pass@10 {\small$\uparrow$}     & \textbf{31.40}  & 22.60 & 23.80          & 24.20             & 26.80          & 28.40          & 29.20          & 28.80          \\
    NaturalQuestions {\small$\uparrow$} & 13.45           & 12.49 & 12.37          & \textbf{13.51}    & 13.45          & 13.40          & 12.31          & 13.19          \\
    PIQA {\small$\uparrow$}             & 73.99           & 73.39 & 73.99          & 73.83             & 74.05          & \textbf{74.48} & 73.67          & 73.99          \\
    RACE.high {\small$\uparrow$}        & \textbf{37.19}  & 33.65 & 33.25          & 33.56             & 35.71          & 35.71          & 35.11          & 35.99          \\
    RACE.mid {\small$\uparrow$}         & \textbf{50.63}  & 46.59 & 45.54          & 46.52             & 49.09          & 48.89          & 48.40          & 49.30          \\
    SIQA {\small$\uparrow$}             & 42.48           & 40.23 & 41.61          & \textbf{42.53}    & 42.02          & 41.71          & 41.91          & 41.71          \\
    TriviaQA {\small$\uparrow$}         & \textbf{30.11}  & 27.96 & 28.15          & 28.95             & 29.59          & 28.26          & 28.98          & 29.95          \\
    Winogrande {\small$\uparrow$}       & 61.41           & 58.01 & \textbf{62.12} & 62.04             & 59.91          & 58.33          & 59.27          & 59.51          \\
    \midrule
    \textbf{average} {\small$\uparrow$}    & \textbf{41.57}  &  38.93  & 39.63  & 39.81    & 40.56          & 40.69          & 40.51          & \textbf{40.88}   \\
    \bottomrule
    \end{tabular}}
    \caption{Validation perplexity and accuracy on short-context reasoning and commonsense tasks. All models have 1.4B parameters. To compute the transformer FLOPs we use the training sequence length of 16384.
    }
    \label{tab:hybrids vs pure acc}
    \label{tab:windows acc}
\end{table}

Table \ref{tab:hybrids vs pure acc} shows that the performance of hybrid models on short-context reasoning benchmarks is higher than that of a xLSTM and also slightly higher than that of a pure SWA architecture. This further highlights the fact that for short contexts, hybrid models leverage the high precision of the softmax sliding window attention layers.
Hybrid models therefore take the strong short-context performance of softmax attention, and the improved long-context recall ability of the Linear Attention layers.

\subsection{In search of an optimal window size for hybrids}
\label{sec:optimal window}

Hereafter, we establish that windows that are too long actually hinder the Linear Attention layers from learning to model long-term dependencies during training.
We hypothesize that this degradation is
due to under-training of the linear attention layers on the long-context recall task.
To validate this hypothesis, we train SWAX with varying window sizes in $\{128, 256, 512, 1024, 2048\}$ and test the models on both short context reasoning tasks and, more importantly, also on long-context recall tasks like RULER NIAH.

\paragraph{Short-context performance.}
\cite{de2024griffinmixinggatedlinear} experimented with different window sizes to find the optimal sliding window size.
However, they only evaluated the different window sizes using the validation perplexity.
Table \ref{tab:windows acc} shows that, indeed, the hybrid with the largest softmax attention window (SWAX:2048) has the best performance from the validation perplexity point of view.

However, raw perplexity is not sufficient to accurately predict performance in downstream tasks, and especially not in long-context modeling tasks \citep{fang2025wrongperplexitylongcontextlanguage}.
Thus, we also evaluate the impact of the window size on short-context reasoning and common sense benchmarks and on long-context retrieval tasks from the RULER benchmark.
Table \ref{tab:windows acc} shows that on short-context reasoning benchmarks all window sizes except the shortest one, 128, give similar results, with the best performing hybrid being the one trained with the longest window size of 2048.
The worse performance of the shortest window of size 128 is understandable as most prompts, even from those relatively short reasoning benchmarks, do not fit within a sliding window of 128 tokens.

\paragraph{Long-context performance.}

\begin{figure}[h]
    \centering
    \includegraphics[width=\linewidth]{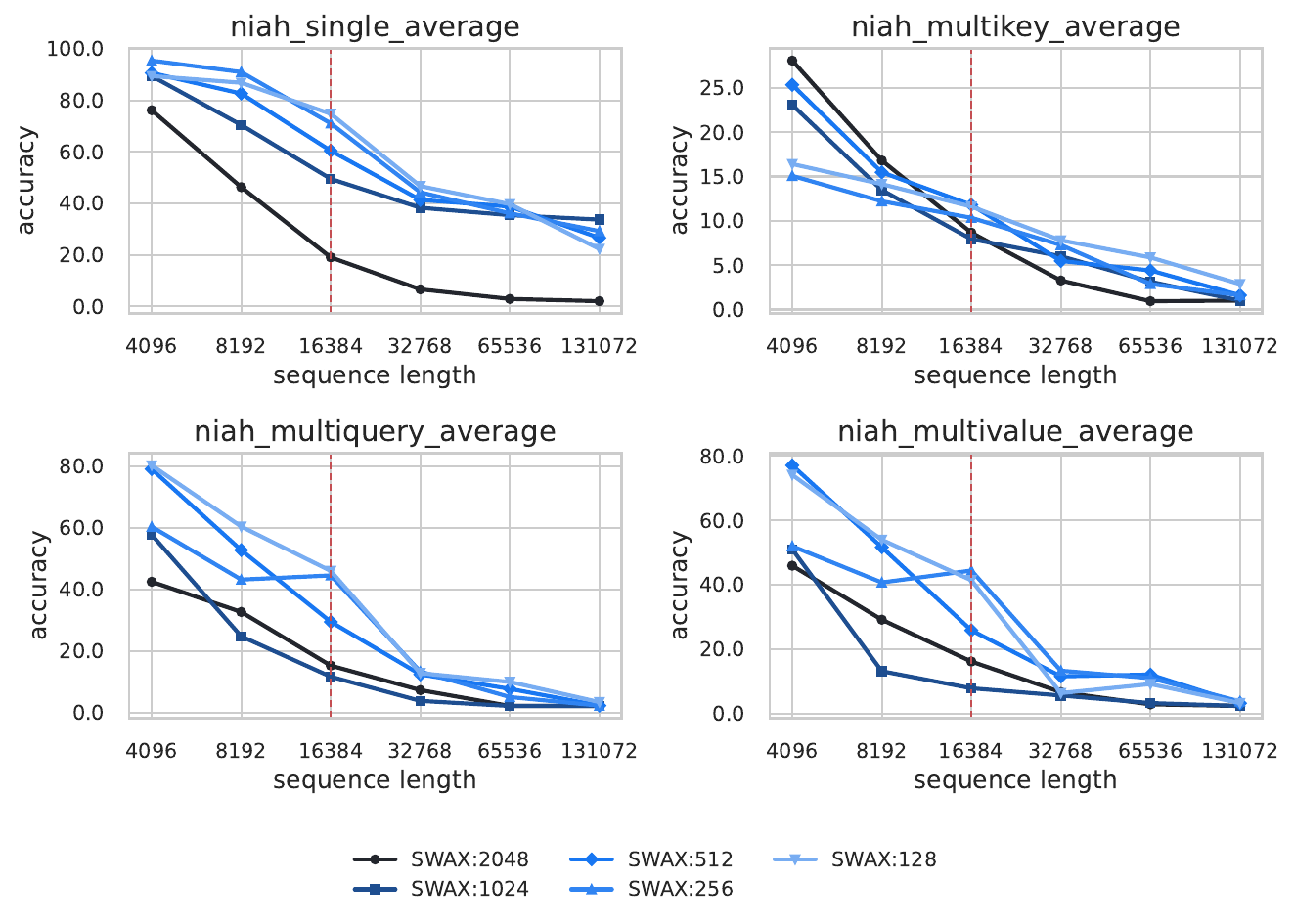}
    \caption{RULER NIAH subtasks accuracy for 1.4B SWAX models with different window sizes}
    \label{fig:NIAH hybrids}
\end{figure}

Figure \ref{fig:NIAH hybrids} shows that once tested on longer sequences, the performance of the hybrid trained with a window size of 2048 drops the most.
On the other hand, the SWAX models trained with shorter window sizes like 128, 256, and 512, maintain better performance even up to sequence lengths of 65k and 131k tokens.
On the NIAH single task, SWAX models with a shorter window have around 30\% recall accuracy at 131k sequence length, while the SWAX with a window of 2048 has near 0\% recall.
Even the shortest sliding window of size 128, which consistently underperformed the longest ones in terms of PPL and short-context reasoning, significantly outperforms the model with the 2048 window length on all RULER NIAH tasks.

As shown in Figure \ref{fig:traintest window niah}, averaging over all sequence lengths and NIAH tasks, the SWAX model with a window size of 128 actually performs the \textit{best} out of all the window sizes we tested.
In particular, it outperforms the 2048 window size by 16 accuracy points.
In other words, the SWAX with the shortest window has a recall \textit{88.9\% higher} than the SWAX with the longest window.
The most likely cause for this phenomenon is that during training, most of the dependencies to model fall inside the 2048 tokens window.

Therefore, during pretraining, it was advantageous for the model with a window of 2048 to use the more precise softmax attention from the sliding window rather than having to rely on the less precise Linear Attention layers to model most dependencies.
However, once tested on longer sequence length where the dependencies are outside of the window length, the model does not extrapolate since it never learned to rely on the Linear Attention layers to do long-context modeling.

On the other hand, the models with shorter windows \textit{had to} rely on the Linear Attention layers to propagate information since many dependencies fell outside of the sliding window.
We give further evidence in Appendix \ref{sec:appendix A} which further indicates that this is indeed the reason for the poor long-context performance of hybrids with long sliding windows.

All these results show that, contrary to previous belief, longer sliding windows do not always provide better performance and can even have a  \textit{negative} impact when extrapolating to tasks beyond the sliding window size and training sequence length.
On the contrary, shorter window sizes push the Linear Attention layers, that have a global receptive field, to receive more supervisory signal and specialize in long-context dependencies.
Overall, shorter sliding windows allow the model to better extrapolate to tasks beyond the sliding window size and even far beyond the training sequence length.
This also means that shorter windows are not just a way of reducing computational cost or maximize hardware utilization as was often thought to be the case, as in \cite{arora2025simplelinearattentionlanguage} and \cite{de2024griffinmixinggatedlinear}.

\subsection{Different window sizes at train and test time}

\begin{table}[t]
\begin{minipage}{0.35\linewidth}
    \includegraphics[width=\linewidth]{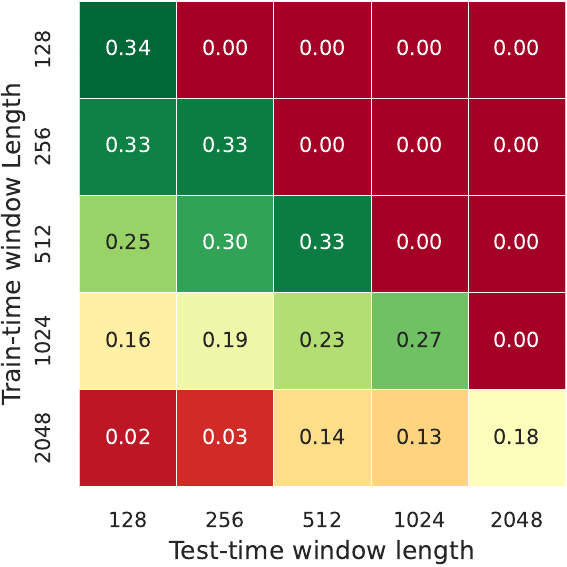}
    \captionsetup{type=figure}
    \caption{average NIAH accuracy of 1.4B SWAX models depending on their train and test time window sizes.
    \label{fig:traintest window niah}}
    \end{minipage}
    \hfill
    \begin{minipage}{0.55\linewidth}
   \caption{Validation Perplexity and accuracy on downstream tasks. Stochastic models use $w$\,=\,$2048$ by default and switch to $w$\,=\,$128$ with probability $p$\,=\,$0.5$ at 1.4B scale and $p$\,=\,$0.75$ at 7B scale.  \vspace{-0.7em}
    \label{tab:sto accs}}
    \scalebox{0.99}{
    {\footnotesize
    \begin{tabular}{@{\ }l@{\quad}|@{\quad}c@{\ }c@{\ }c@{\quad}|c@{\ }c@{\ }c@{\ }}
\toprule
model -- parameters & \multicolumn{3}{c|}{SWAX 1.4B}  & \multicolumn{3}{c}{SWAX 7B}  \\
\midrule
train window length  & 128 & stochastic & 2048 & 128 & stochastic & 2048 \\
test window length & 128 & 2048 & 2048 & 128 & 2048 & 2048 \\
\midrule
val\_PPL {\small$\downarrow$} & 2.551 & \textbf{2.502} & 2.523 & 2.291 & \textbf{2.272} & 2.283 \\
\midrule
HEplus/pass@10 {\small$\uparrow$}   & 12.20 & 12.80 &15.24 & 24.39 & 24.39 & 26.83 \\
ARC-c {\small$\uparrow$}            & 29.27 & 30.82 & 29.79 & 40.77 & 40.86 & 41.55 \\
ARC-e {\small$\uparrow$}            & 65.75 & 68.71 & 67.15 & 75.05 & 74.46 & 74.80 \\
Hellaswag {\small$\uparrow$}        & 45.37 & 45.51 & 45.91 & 53.34 & 53.59 & 53.69 \\
MBPP/pass@10 {\small$\uparrow$}     & 24.20 & 30.60 & 28.80 & 44.60 & 47.20 & 45.40 \\
NaturalQuestions {\small$\uparrow$} & 13.51 & 12.47 & 13.19 & 22.21  & 23.45 & 23.04 \\
PIQA {\small$\uparrow$}             & 73.83 & 74.43 & 73.99 & 77.26 & 77.97 & 76.55 \\
RACE.high {\small$\uparrow$}        & 33.56 & 35.91 & 35.99 & 37.65 & 38.99 & 39.88 \\
RACE.mid {\small$\uparrow$}         & 46.52 & 48.33 & 49.30 & 52.65 & 54.32 & 54.80 \\
SIQA {\small$\uparrow$}             & 42.53 & 41.86 & 41.71 & 43.55 & 44.22 & 42.78 \\
TriviaQA {\small$\uparrow$}         & 28.95 & 28.99 & 29.95 & 46.20 & 47.15 & 46.80 \\
Winogrande {\small$\uparrow$}       & 62.04 & 59.27 & 59.51 & 67.88 & 67.64 & 65.75 \\
\midrule
average {\small$\uparrow$}          & 39.81 & 40.81 & \textbf{40.88} & 48.80 & \textbf{49.52} & 49.32  \\
\bottomrule
\end{tabular}}}
    \end{minipage}
\end{table}
We now explore a training strategy allowing for a large window size at test time, to have the best reasoning performance possible, while still being trained such that the Linear Attention layers for long-term dependencies and extrapolate to longer sequences.

\paragraph{Length extrapolation.}
As a preliminary analysis, we first evaluate the performance of models when tested with a different window size than the one used at training time.
Figure \ref{fig:traintest window niah} shows that, as expected, naively extending the window size beyond its training length results in catastrophic collapse.
This is a common phenomenon in softmax attention with RoPE which is used in the SWA layers  \citep{peng2023yarnefficientcontextwindow}.
On the other hand, windows of size 1024 and less show little degradation when reducing their train-time window size by half.
Overall, models need to be trained on large windows sizes to be able to use large windows during testing.
At the same time, we cannot allow the models to over-rely on the long softmax attention windows since those do not perform well on very long-context tasks.

\paragraph{Stochastic window size.}
To solve this dilemma, we introduce a training procedure that,  throughout the training, stochastically alternates between a large window size and a small window size.
Our hypothesis is that this will prevent the model from over-relying on the SWA layers, %
while still making the model capable of using the larger window size at test time. %

Moreover, to validate our experiments at a larger scale, we also train 7B parameter models using the same experimental setup as for the 1.4B models.
For the 1.4B experiments, at each new batch of data, we set the window size to 128 with probability $p=0.5$ or leave the default window size of 2048.
At 7B scale, we use a slightly higher probability $p=0.75$ of sampling the short window.
We provide an ablation for the value of $p$ in Appendix~\ref{sec:appendix B}.
Finally, to force the model to make better use of the larger test-time window of size 2048, we anneal the stochastic training procedure by not sampling the smaller window size anymore for the last 10\% of training.
We find that this short period of fixed windows at the end of the training significantly helps short-context performance without  degrading long-context performance.
We provide an ablation of the annealing in Appendix~\ref{sec:appendix B}.

Table \ref{tab:sto accs} shows how training with a stochastic window size alternating between 128 and 2048 and annealing gives a short-context performance comparable to or even better than training with a fixed window size of 2048.
In particular, at both 1.4B and 7B scales, stochastic training gives considerably better short-context performance than a fixed-sized window of 128.
From a validation perplexity perspective, the stochastic window size outperforms all models trained with fixed window sizes at all parameter scales.
Therefore, training with a stochastic window size and testing with a longer window yields better results on short-context tasks compared to a short window at both train and test time.

Compared to training with a fixed long sliding window of size 2048, stochastic training gives comparable performance at the 1.4B scale and even slightly superior performance at the 7B scale on short-context reasoning tasks.
This indicates that indeed, even if during training the model has seen the longer window size only part of the time, it is still able to take advantage of the longer window size for short-medium context reasoning tasks.

\paragraph{Long-context performance of the stochastic training.}
We evaluate the stochastically trained SWAX models with annealing on the last 10\% of the training.
This is to ascertain this strategy gives a performance on long-context tasks as good as short-window variants.

\begin{figure}[t]
\begin{minipage}{0.48\linewidth}
\includegraphics[width=\linewidth]{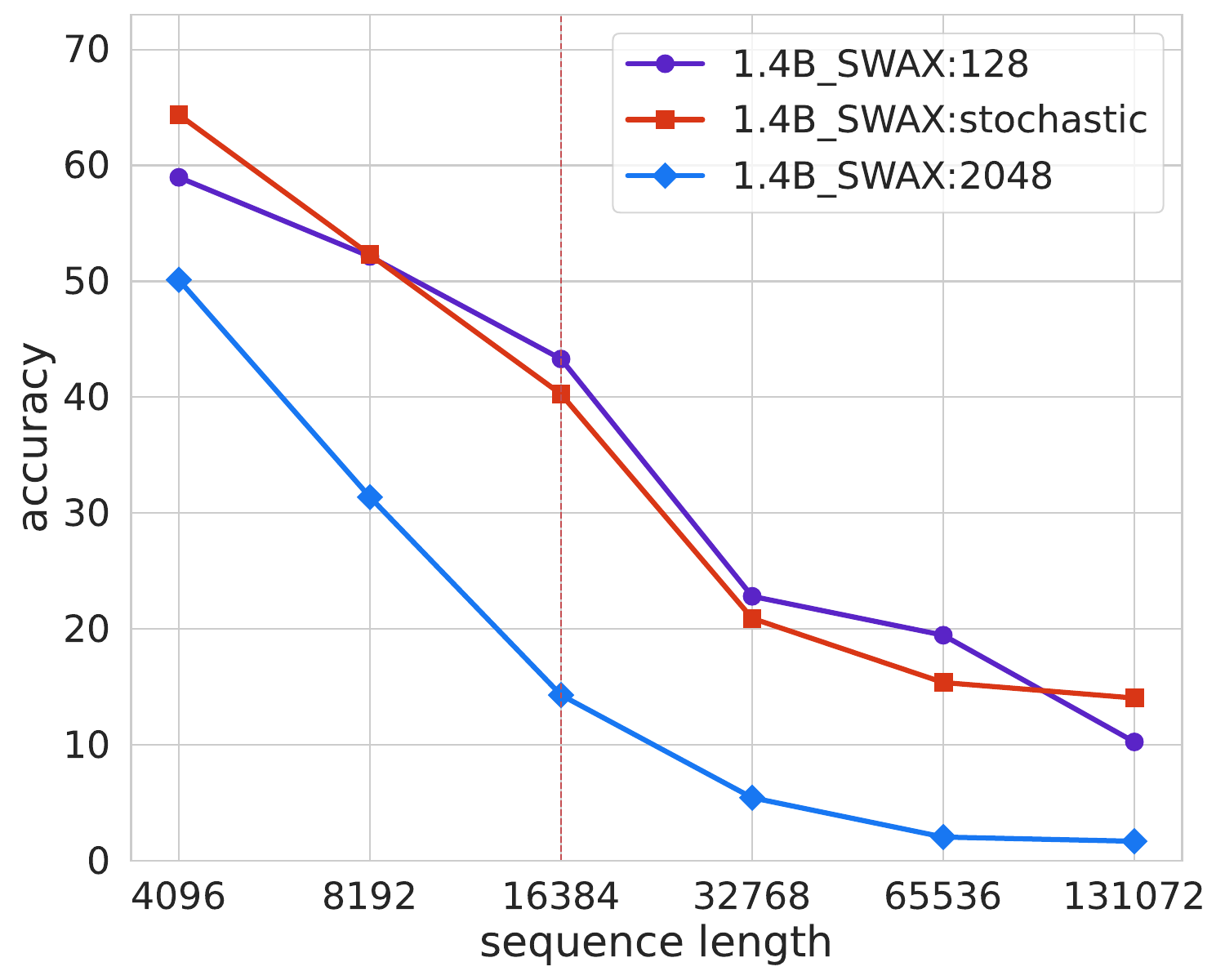}
\caption{Average RULER NIAH accuracy of 1.4B SWAX models with different window sizes.
\label{fig:sto NIAH}
}
\end{minipage}
\hfill
\begin{minipage}{0.48\linewidth}
\includegraphics[width=\linewidth]{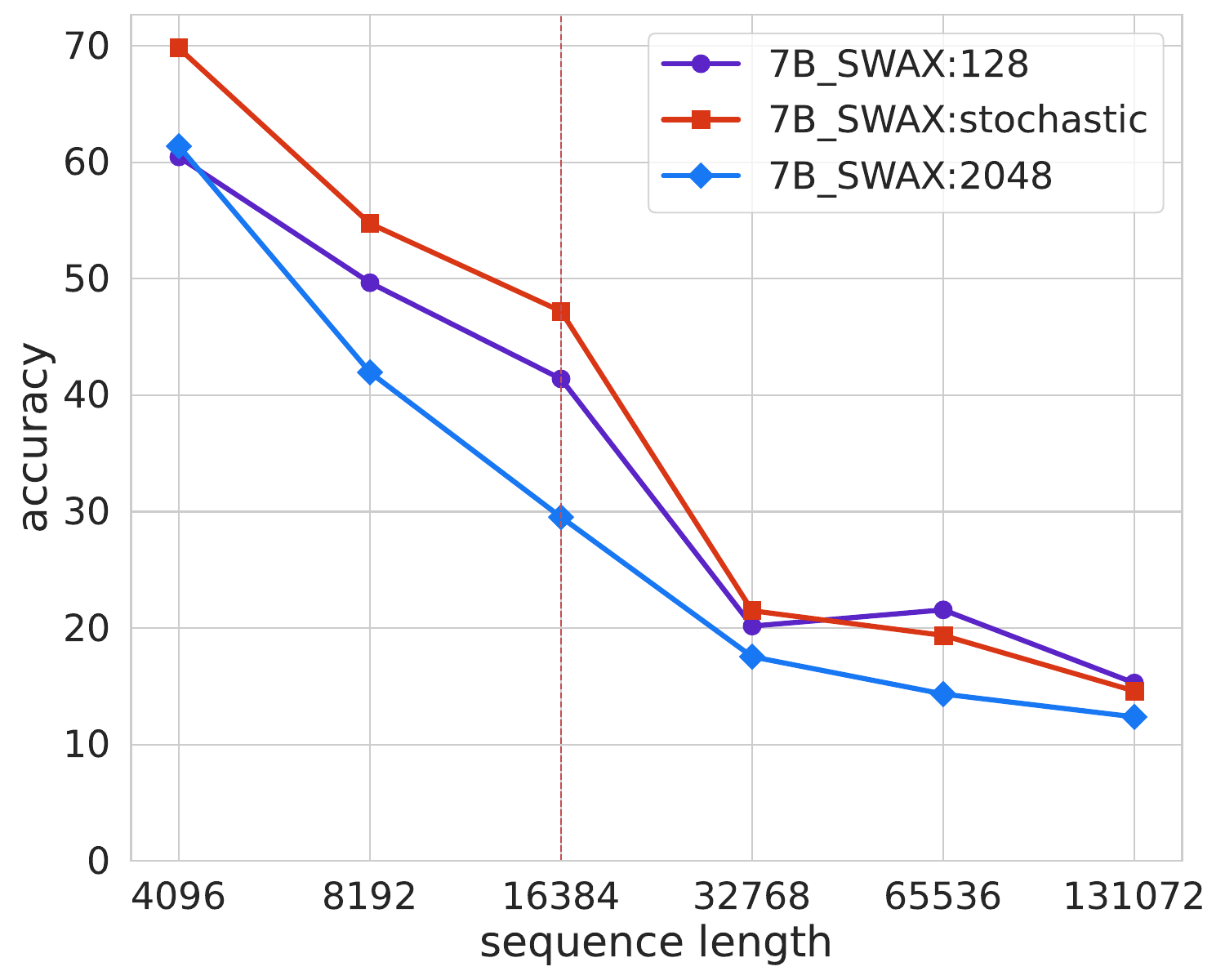}
\caption{Average RULER NIAH accuracy of 7B SWAX models with different window sizes.
\label{fig:7B sto NIAH}
}
\end{minipage}
\end{figure}

Figures \ref{fig:sto NIAH} and \ref{fig:7B sto NIAH} show that on RULER's long-context recall tasks, the models trained with a stochastic window size and annealing perform on par or better than the model trained with the short window of 128, and drastically better than the hybrid model trained with a window of 2048 tokens.
For instance, at 1.4B parameter scale depicted in Figure \ref{fig:sto NIAH}, the stochastic training SWAX model performs similarly to the short-window SWAX.
Figure \ref{fig:7B sto NIAH} shows that stochastic training also improves the long-context performance of models at 7B scale.
At 7B scale, compared to using a fixed sliding window of size 2048, the stochastic training gives much better retrieval accuracy at all sequence lengths.
Furthermore, just as at 1.4B scale, stochastic training gives similar or even superior long-context performance compared to using a short window throughout training.
Overall, a stochastic window size at training time maintains all the benefits of having a short window for long-context recall, and most if not all of the benefits of a longer window for short/medium-context reasoning tasks.

Moreover, this result further confirms that the poor performance of hybrids trained on long windows is not intrinsically due to a long sliding window at test time.
Instead, these results show that the poor length-generalization of hybrids with long sliding windows is due to the training procedure.
Indeed, if the model is allowed to use the long sliding window throughout training, it will over-rely on the more precise softmax attention of the sliding window even for recall tasks, which will not extrapolate to longer sequence lengths.
It will under-utilize the Linear Attention Layers for the long-term recall task.
On the contrary, if during training the model is not allowed to rely on the long window, and is instead stochastically forced to use a shorter window, then the linear attention will be required on the medium/long-term recall tasks.
Since, essentially, this amounts to stochastically reducing the capacity of the model to make it more robust, this can be seen as a form of dropout \citep{JMLR:v15:srivastava14a} on the attention mechanism.

\paragraph{More long-context benchmarks.}

The Ruler tasks are artificial long context benchmarks, built to be sensible to model variants. 
We also evaluated SWAX on more realistic tasks from several families of benchmarks: 
\begin{itemize}
    \item \textbf{LongBench}~\citep{bai2024longbench} is a bilingual, multitask, and comprehensive benchmark to evaluate the long context understanding capabilities of LLMs. It is
    composed of six categories and twenty one different tasks. It evaluates long-text application scenarios such as single-document QA, multi-document QA, summarization, few-shot learning, synthetic tasks and code completion.
    \item \textbf{Babilong}~\citep{kuratov2024babilong} This benchmark extends the Babi tasks~\citep{Weston2015TowardsAQ}, it is designed to evaluate long-term memory and reasoning capabilities of LLMs. Babilong uses long input sequences, making it suitable for evaluating models with advanced memory architectures. The main categories are: single book reasoning, memory and retrieval and temporal and spatial reasoning.
    \item \textbf{LongBench2}~\citep{bai2024longbench2} builds on the original LongBench, expanding both the scale and diversity of tasks to better stress-test LLMs for real-world long-context scenarios.
\end{itemize}

The results are summarized in Table~\ref{tab:otherLC}~(bottom). 
They show that the tasks are difficult on average for such small models. 
Stochastic training outperforms fixed-size training in some settings, but for others, the long context (2048) performs better. 
In Table~\ref{tab:longbenchsumm}, Appendix~\ref{sec:longbench}  we present the average performance of the LongBench summarization and question-answering tasks, where the same conclusion holds.
\begin{table}[htbp]
\begin{minipage}{0.45\linewidth}
    \scalebox{0.85}{
    \begin{tabular}{l|r|r|r}
    \toprule 
            & \multicolumn{3}{c}{Training SWA size}\\
    Task family   & 128 & stochastic & 2048 \\
    \midrule 
    Longbench     & 10.67 & \textbf{11.79} & 11.14 \\
    Longbench2    & 22.99 & \textbf{27.26} & 22.65 \\
    Babilong      &  4.68 & \textbf{9.35} & 7.29 \\
    \midrule 
    \multicolumn{4}{c}{Gated DeltaNet}\\
    \midrule  
    Ruler multiquery NIAH    & 40.58 &   \textbf{40.90} & 28.80 \\
    Longbench     & 9.14   & 9.94 & \textbf{11.42} \\
    Babilong      & 3.74  & 7.21 & \textbf{8.95} \\
    \bottomrule
    \end{tabular}
    }
    \captionof{table}{Other long-context benchmarks on our SWAX architecture with 1.4B parameters. 
        The results are averaged over all tasks within a family.
        Top: xLSTM linear attention memory, bottom: Gated DeltaNet linear attention.
        For Gated DeltaNet we also report the averaged performance of Ruler multiquery NIAH, averaged over sequence sizes 4096, 8192 and 16384.}
    \label{tab:otherLC}
\end{minipage}
\hfill
\begin{minipage}{0.45\linewidth}
    \includegraphics[width=\linewidth]{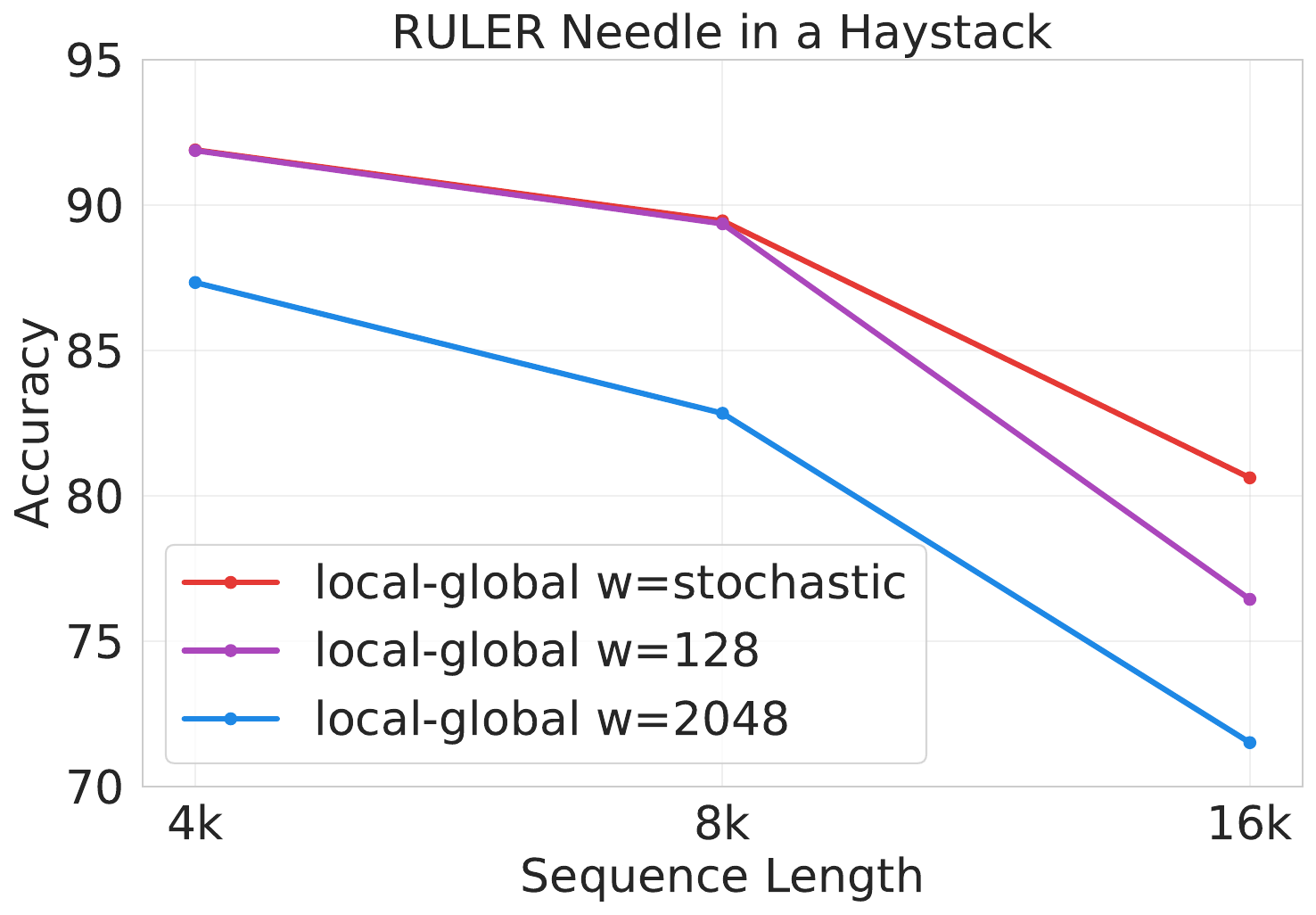}
    \captionof{figure}{RULER Needle in a Haystack average accuracy for 1.4B local:global models over NIAH single1, 2, 3, Multiquery and Multikey, from 4k sequence length up to 16k.  }
    \label{fig:local-global}
\end{minipage}
\end{table}

\subsection{Experiments with Gated DeltaNet and full attention}

The Gated DeltaNet~\citep{yang2025gateddeltanetworksimproving} is a state-of-the-art hybrid architecture that combines gating and delta-update rule to update the linear attention memory. 
To assess whether our observations apply to other linear attention architectures, we replace the xLSTM component in the previous experiments with a Gated DeltaNet layer (we call this combination SWAX/GDN). 
We did not change any training setting, and ran the same long-context benchmarks. 
Table~\ref{tab:otherLC}~(top) shows that hybrid architectures with Gated DeltaNets also benefit from the stochastic training.

To further validate our findings, we also trained and evaluated local-global architectures which alternate SWA layers with full attention layers. This setting, which keeps some amount of full attention layer in the architecture, is commonly used by SoTA open source models like \citet{openai2025gptoss120bgptoss20bmodel} or \citet{singh2026arceetrinity}.
\
As shown in figure \ref{fig:local-global}, using larger sliding windows during training also degrades the long-context performance of local-global architectures with full attention. 
This reinforces the fact that our findings generalize to many kind of hybrid attention architectures.

\section{Conclusion}
Through an empirical analysis of hybrid architectures, we evidence the counter-intuitive fact that shorter sliding windows lead to better length-extrapolation on retrieval tasks.
Moreover, we introduce a training procedure that stochastically changes the window size throughout training. %
This training procedure offers a strong performance on short-context tasks (enabled by longer sliding windows) and the length-extrapolation ability of Linear Attention layers, enabled by shorter windows at training time.

%% file: appendix.tex
\section{Results of pure SWA models}
\label{sec:appendix A}

In section \ref{sec:optimal window} we hypothesize that the worse performance of SWAX models with long windows comes from the model utilizing the SWA layers instead of the xLSTM layers.
To further confirm this hypothesis, we train a 1.4B pure SWA model with a window size of 2048 and compare its performance to the SWAX model with the same window size.
If the hypothesis that the SWAX model relies on the SWA layers for recall is valid, then we expect its performance to be similar to that of a pure SWA architecture.

\begin{figure}[h]
    \centering
    \includegraphics[width=0.5\linewidth]{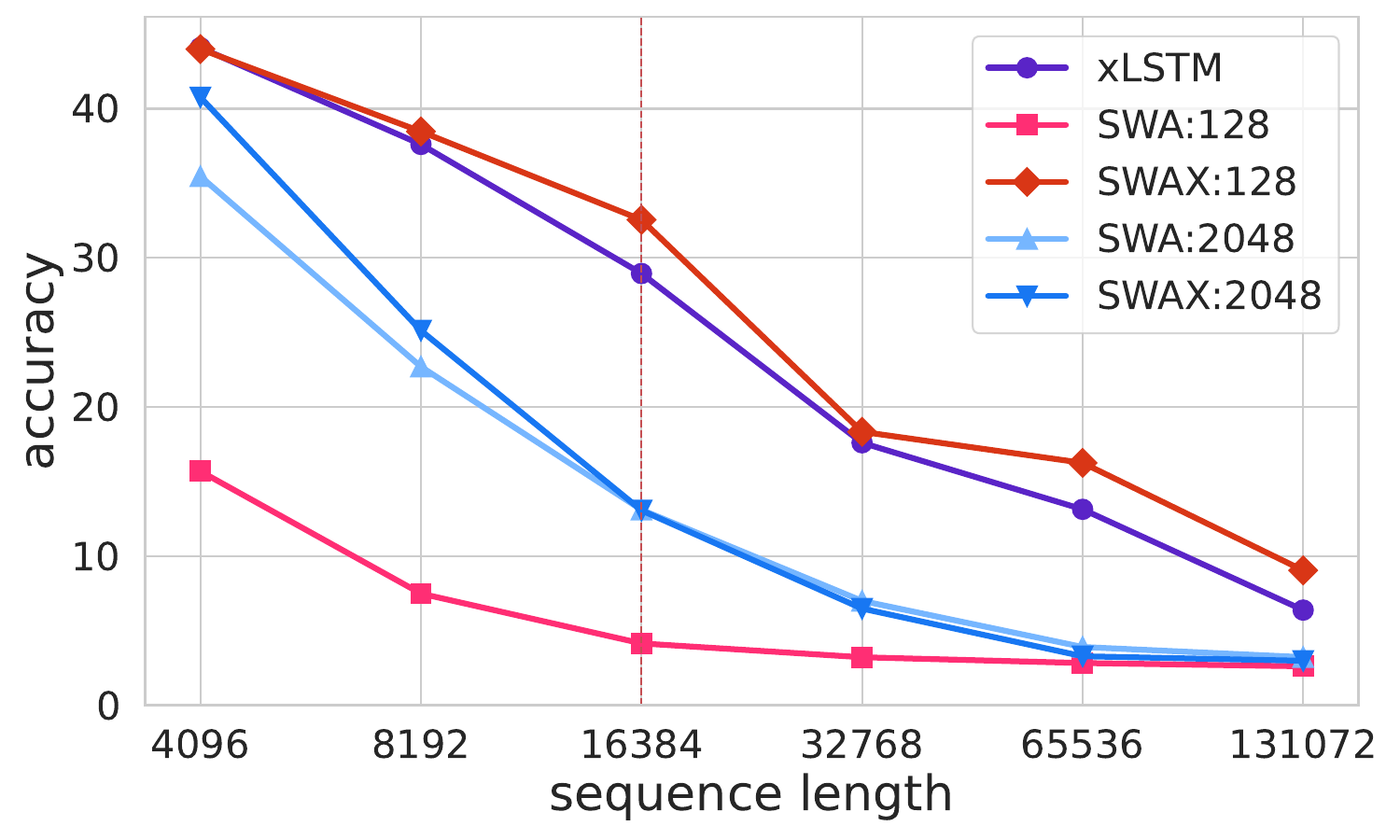}
    \caption{Average accuracy of 1.4B parameter models over all RULER NIAH tasks.}
    \label{fig:SWA-SWAX}
\end{figure}

Figure \ref{fig:SWA-SWAX} shows that indeed, the SWAX model trained with a window of 2048 performs very similarly to the pure SWA architecture.
On the other hand, the accuracy the SWAX model trained with a window of 128 is dissimilar from that of the pure SWA model with a window of 128.
This further evidences that low long-context performance of hybrid models with long windows comes from the model over-relying on the SWA layers for long-context recall instead of using the xLSTM layers.

\section{Stochastic sampling probability and annealing of stochasticity}
\label{sec:appendix B}
In this experiment we perform an ablation on the stochastic sampling probability $p$ its schedule during training for the two model sizes 1.4B and 7B we consider in our study.
Table \ref{tab:stochastic_7B_full_results} shows that, at 7B scale, a higher probability of sampling the small window during training is necessary to significantly improve short and long context performance compared to the probability of 0.5 which worked at 1.4B scale.
Looking at the impact of annealing, i.e. using a stochastic window size for the first X\%  of training, at both 1.4B and 7B scale, annealing improves short-context performance compared to keeping the stochasticity until the end of training. At 7B scale, the annealed SWAX model even performs better on short-context than the SWAX model trained with a fixed window size of 2048.
In terms of long-context performance, compared to keeping the stochasticity until the end of training, annealing slightly degrades the long-context performance at 1.4B scale but keeps or even slightly improves long-context performance at 7B.
We believe that exploring different annealing procedures might provide even better short-context performance improvements while --- at the same time --- keeping good long-context performance.

\begin{table}[t]
\centering
\scalebox{0.99}
    {\footnotesize
    \begin{tabular}{l@{\hspace{2cm}}|c|cccccc|cc}
\toprule
model & xLSTM &  \multicolumn{6}{c|}{SWAX} & \multicolumn{2}{c}{SWAX} \\
parameters & 7B & \multicolumn{6}{c|}{7B}  & \multicolumn{2}{c}{1.4B} \\
\midrule
train-time window & NA & 128 & $p$=0.9 & $p$=0.75 & $p^{90\%}$=0.75 & $p$=0.5 & 2048 & $p$=0.5 & $p^{90\%}$=0.5\\
test-time window & NA & 128 & 2048 & 2048 & 2048 & 2048 & 2048 & 2048 & 2048 \\
\midrule
\textbf{niah\_single} & 61.20 & 62.43 & 58.99 & \textbf{63.36} & 63.20 & 55.27 & 53.46 & 62.61 & 61.59\\
\textbf{niah\_multiquery} & \textbf{44.18} & 34.78 & 35.95 & 32.72 & 32.23 & 17.94 & 21.62 & 30.85 & 27.42\\
\textbf{niah\_multikey} & 10.14 & 9.96 & 13.32 & 17.23 & \textbf{17.86} & 14.34 & 12.29 & 14.52 & 12.19\\
\textbf{niah\_multivalue} & \textbf{39.28} & 26.11 & 23.55 & 27.32 & 27.56 & 19.48 & 17.30 & 30.63 & 27.63\\
\midrule
\textbf{niah\_average} & 37.19 & 34.76 & 34.55 & 37.73 & \textbf{37.87} & 30.78 & 29.52 & \textbf{36.61} & 34.55 \\
\midrule
\textbf{HEplus/pass@10}  & 25.00 & 24.39 & 25.61 & 23.17 & 24.39 & 21.95 & \textbf{26.83} & 13.41 & 12.80\\
\textbf{arc-c}           & 37.42 & 40.77 & 40.00 & 40.86 & 40.86 & 40.09 & \textbf{41.55} & 32.45 & 30.82\\
\textbf{arc-e}           & 73.32 & \textbf{75.05} & 74.76 & 74.50 & 74.46 & 74.63 & 74.80 & 67.23 & 68.71\\
\textbf{hella}           & 52.95 & 53.34 & 53.34 & 53.48 & 53.59 & 53.47 & \textbf{53.69} & 45.61 & 45.51\\
\textbf{mbpp/pass@10}    & 43.80 & 44.60 & 43.60 & 42.80 & \textbf{47.20} & 45.80 & 45.40 & 28.00 & 30.60\\
\textbf{nq}              & 22.12 & 22.21 & 23.12 & 23.16 & \textbf{23.45} & 22.58 & 23.04 & 12.95 & 12.47\\
\textbf{piqa}            & 76.93 & 77.26 & 76.71 & \textbf{78.07} & 77.97 & 77.31 & 76.55 & 74.32 & 74.43\\
\textbf{race.high}       & 37.02 & 37.65 & 37.71 & 38.74 & 38.99 & 38.74 & \textbf{39.88} & 34.82 & 35.91\\
\textbf{race.mid}        & 52.85 & 52.65 & 54.11 & 54.32 & 54.32 & 53.90 & \textbf{54.80} & 48.26 & 48.33\\
\textbf{siqa}            & 43.96 & 43.55 & 44.06 & 44.01 & 44.22 & \textbf{44.37} & 42.78 & 41.91 & 41.86\\
\textbf{tqa}             & 46.39 & 46.20 & \textbf{47.44} & 46.90 & 47.15 & 46.55 & 46.80 & 29.35 & 28.99\\
\textbf{wino}            & 66.06 & 67.88 & \textbf{68.51} & 67.32 & 67.64 & 66.77 & 65.75 & 59.20 & 59.27\\
\midrule
\textbf{short-average}         & 48.15 & 48.80 & 49.08 & 48.95 & \textbf{49.52} & 48.85 & 49.32 & 40.62 & \textbf{40.81}\\
\bottomrule
    \end{tabular}
    }
    \caption{NIAH and downstream tasks accuracy for 7B models. p indicates the probability of using a window of 128 for a batch, otherwise using a window of 2048. $p^{90\%}$ indicates annealing, i.e., only doing the stochastic window size for the first 90\% of the training and then using a fixed window size of 2048 for the rest of training. NIAH single and multikey results are the average overall all 3 sub-tasks for each.}
    \label{tab:stochastic_7B_full_results}
\end{table}

\section{Benchmarks}
\label{sec:benchmarks}
\paragraph{Code generation}
We use two benchmarks that evaluate the code generation capabilities of AI models: HumanEval+ and MBPP.
\begin{itemize}
    \item The HumanEval+~\citep{evalplus} benchmark is an extension of HumanEval~\citep{chen2021codex}, which is designed to evaluate the functional correctness of code generated by AI models.
    \item MBPP~\citep{austin2021programsynthesislargelanguage} is designed to evaluate the code generation abilities of AI models, particularly for Python programming tasks.
\end{itemize}

\paragraph{Common sense and general reasoning.}  We use benchmarks consisting of question-answer or multiple-choice questions designed to evaluate the common sense reasoning abilities of AI models, particularly in the context of natural language understanding: HellaSWAG~\citep{zellers2019hellaswagmachinereallyfinish}, ARC~\citep{clark2018thinksolvedquestionanswering}, PIQA~\citep{bisk2019piqareasoningphysicalcommonsense}, SIQA \citep{sap2019socialiqacommonsensereasoningsocial}, Winogrande~\citep{sakaguchi2019winograndeadversarialwinogradschema}, NaturalQuestions~\citep{kwiatkowski-etal-2019-natural}, RACE~\citep{lai2017racelargescalereadingcomprehension} and TQA~\citep{joshi-etal-2017-triviaqa}.
\section{LongBench summarisation and QA tasks performance}
\label{sec:longbench}
We can report the averages for all the summarization and question answering tasks, we observe that training with stochastic SWA size outperforms the fixed-size training windows.
\begin{table}[h]
\centering
\begin{tabular}{lccc}
\toprule
Task & SWAX:128 & SWAX:2048 & SWAX:stochastic \\
\midrule
Longbench summarization tasks (average) & 10.570 & 11.152 & \textbf{12.429} \\
Longbench QA tasks (average) & 5.203 & 6.112 & \textbf{7.257} \\
\bottomrule
\end{tabular}\caption{Average performance for summarization and question-answering LongBench tasks}
\label{tab:longbenchsumm}
\end{table}
\begin{itemize}

\item The LongBench summarization tasks are: longbench\_gov\_report,longbench\_qmsum, longbench\_multi\_news and longbench\_vcsum. 
\item The  LongBench QA tasks, which include single-QA and multi-QA tasks, are: longbench\_hotpotqa, longbench\_2wikimqa, longbench\_musique, longbench\_dureader, longbench\_narrativeqa, longbench\_qasper, longbench\_multifieldqa\_en and longbench\_multifieldqa\_zh.
\end{itemize}

%% file: main_arxiv.bbl
\begin{thebibliography}{54}
\providecommand{\natexlab}[1]{#1}
\providecommand{\url}[1]{\texttt{#1}}
\expandafter\ifx\csname urlstyle\endcsname\relax
  \providecommand{\doi}[1]{doi: #1}\else
  \providecommand{\doi}{doi: \begingroup \urlstyle{rm}\Url}\fi

\bibitem[Arora et~al.(2025)Arora, Eyuboglu, Zhang, Timalsina, Alberti, Zinsley,
  Zou, Rudra, and Ré]{arora2025simplelinearattentionlanguage}
Simran Arora, Sabri Eyuboglu, Michael Zhang, Aman Timalsina, Silas Alberti,
  Dylan Zinsley, James Zou, Atri Rudra, and Christopher Ré.
\newblock Simple linear attention language models balance the recall-throughput
  tradeoff, 2025.
\newblock URL \url{https://arxiv.org/abs/2402.18668}.

\bibitem[Austin et~al.(2021)Austin, Odena, Nye, Bosma, Michalewski, Dohan,
  Jiang, Cai, Terry, Le, and Sutton]{austin2021programsynthesislargelanguage}
Jacob Austin, Augustus Odena, Maxwell Nye, Maarten Bosma, Henryk Michalewski,
  David Dohan, Ellen Jiang, Carrie Cai, Michael Terry, Quoc Le, and Charles
  Sutton.
\newblock Program synthesis with large language models, 2021.
\newblock URL \url{https://arxiv.org/abs/2108.07732}.

\bibitem[Ba et~al.(2016)Ba, Kiros, and Hinton]{ba2016layernormalization}
Jimmy~Lei Ba, Jamie~Ryan Kiros, and Geoffrey~E. Hinton.
\newblock Layer normalization, 2016.
\newblock URL \url{https://arxiv.org/abs/1607.06450}.

\bibitem[Bai et~al.(2024{\natexlab{a}})Bai, Lv, Zhang, Lyu, Tang, Huang, Du,
  Liu, Zeng, Hou, Dong, Tang, and Li]{bai2024longbench}
Yushi Bai, Xin Lv, Jiajie Zhang, Hongchang Lyu, Jiankai Tang, Zhidian Huang,
  Zhengxiao Du, Xiao Liu, Aohan Zeng, Lei Hou, Yuxiao Dong, Jie Tang, and
  Juanzi Li.
\newblock {L}ong{B}ench: A bilingual, multitask benchmark for long context
  understanding.
\newblock In \emph{Proceedings of the 62nd Annual Meeting of the Association
  for Computational Linguistics (Volume 1: Long Papers)}, pp.\  3119--3137,
  Bangkok, Thailand, August 2024{\natexlab{a}}. Association for Computational
  Linguistics.
\newblock \doi{10.18653/v1/2024.acl-long.172}.
\newblock URL \url{https://aclanthology.org/2024.acl-long.172}.

\bibitem[Bai et~al.(2024{\natexlab{b}})Bai, Tu, Zhang, Peng, Wang, Lv, Cao, Xu,
  Hou, Dong, Tang, and Li]{bai2024longbench2}
Yushi Bai, Shangqing Tu, Jiajie Zhang, Hao Peng, Xiaozhi Wang, Xin Lv, Shulin
  Cao, Jiazheng Xu, Lei Hou, Yuxiao Dong, Jie Tang, and Juanzi Li.
\newblock Longbench v2: Towards deeper understanding and reasoning on realistic
  long-context multitasks.
\newblock \emph{arXiv preprint arXiv:2412.15204}, 2024{\natexlab{b}}.

\bibitem[Beck et~al.(2024)Beck, Pöppel, Spanring, Auer, Prudnikova, Kopp,
  Klambauer, Brandstetter, and Hochreiter]{beck2024xlstmextendedlongshortterm}
Maximilian Beck, Korbinian Pöppel, Markus Spanring, Andreas Auer, Oleksandra
  Prudnikova, Michael Kopp, Günter Klambauer, Johannes Brandstetter, and Sepp
  Hochreiter.
\newblock xlstm: Extended long short-term memory, 2024.
\newblock URL \url{https://arxiv.org/abs/2405.04517}.

\bibitem[Beck et~al.(2025{\natexlab{a}})Beck, Pöppel, Lippe, and
  Hochreiter]{beck:25tfla}
Maximilian Beck, Korbinian Pöppel, Phillip Lippe, and Sepp Hochreiter.
\newblock {Tiled Flash Linear Attention}: More efficient linear rnn and xlstm
  kernels.
\newblock \emph{arXiv}, 2503.14376, 2025{\natexlab{a}}.
\newblock URL \url{https://arxiv.org/abs/2503.14376}.

\bibitem[Beck et~al.(2025{\natexlab{b}})Beck, Pöppel, Lippe, Kurle, Blies,
  Klambauer, Böck, and Hochreiter]{beck2025xlstm7brecurrentllm}
Maximilian Beck, Korbinian Pöppel, Phillip Lippe, Richard Kurle, Patrick~M.
  Blies, Günter Klambauer, Sebastian Böck, and Sepp Hochreiter.
\newblock xlstm 7b: A recurrent llm for fast and efficient inference,
  2025{\natexlab{b}}.
\newblock URL \url{https://arxiv.org/abs/2503.13427}.

\bibitem[Beltagy et~al.(2020)Beltagy, Peters, and
  Cohan]{beltagy2020longformerlongdocumenttransformer}
Iz~Beltagy, Matthew~E. Peters, and Arman Cohan.
\newblock Longformer: The long-document transformer, 2020.
\newblock URL \url{https://arxiv.org/abs/2004.05150}.

\bibitem[Bisk et~al.(2019)Bisk, Zellers, Bras, Gao, and
  Choi]{bisk2019piqareasoningphysicalcommonsense}
Yonatan Bisk, Rowan Zellers, Ronan~Le Bras, Jianfeng Gao, and Yejin Choi.
\newblock Piqa: Reasoning about physical commonsense in natural language, 2019.
\newblock URL \url{https://arxiv.org/abs/1911.11641}.

\bibitem[Chen et~al.(2021)Chen, Tworek, Jun, Yuan, de~Oliveira~Pinto, Kaplan,
  Edwards, Burda, Joseph, Brockman, Ray, Puri, Krueger, Petrov, Khlaaf, Sastry,
  Mishkin, Chan, Gray, Ryder, Pavlov, Power, Kaiser, Bavarian, Winter, Tillet,
  Such, Cummings, Plappert, Chantzis, Barnes, Herbert-Voss, Guss, Nichol,
  Paino, Tezak, Tang, Babuschkin, Balaji, Jain, Saunders, Hesse, Carr, Leike,
  Achiam, Misra, Morikawa, Radford, Knight, Brundage, Murati, Mayer, Welinder,
  McGrew, Amodei, McCandlish, Sutskever, and Zaremba]{chen2021codex}
Mark Chen, Jerry Tworek, Heewoo Jun, Qiming Yuan, Henrique~Ponde
  de~Oliveira~Pinto, Jared Kaplan, Harri Edwards, Yuri Burda, Nicholas Joseph,
  Greg Brockman, Alex Ray, Raul Puri, Gretchen Krueger, Michael Petrov, Heidy
  Khlaaf, Girish Sastry, Pamela Mishkin, Brooke Chan, Scott Gray, Nick Ryder,
  Mikhail Pavlov, Alethea Power, Lukasz Kaiser, Mohammad Bavarian, Clemens
  Winter, Philippe Tillet, Felipe~Petroski Such, Dave Cummings, Matthias
  Plappert, Fotios Chantzis, Elizabeth Barnes, Ariel Herbert-Voss,
  William~Hebgen Guss, Alex Nichol, Alex Paino, Nikolas Tezak, Jie Tang, Igor
  Babuschkin, Suchir Balaji, Shantanu Jain, William Saunders, Christopher
  Hesse, Andrew~N. Carr, Jan Leike, Josh Achiam, Vedant Misra, Evan Morikawa,
  Alec Radford, Matthew Knight, Miles Brundage, Mira Murati, Katie Mayer, Peter
  Welinder, Bob McGrew, Dario Amodei, Sam McCandlish, Ilya Sutskever, and
  Wojciech Zaremba.
\newblock Evaluating large language models trained on code.
\newblock 2021.

\bibitem[Chung et~al.(2014)Chung, Gulcehre, Cho, and
  Bengio]{chung2014empiricalevaluationgatedrecurrent}
Junyoung Chung, Caglar Gulcehre, KyungHyun Cho, and Yoshua Bengio.
\newblock Empirical evaluation of gated recurrent neural networks on sequence
  modeling, 2014.
\newblock URL \url{https://arxiv.org/abs/1412.3555}.

\bibitem[Clark et~al.(2018)Clark, Cowhey, Etzioni, Khot, Sabharwal, Schoenick,
  and Tafjord]{clark2018thinksolvedquestionanswering}
Peter Clark, Isaac Cowhey, Oren Etzioni, Tushar Khot, Ashish Sabharwal, Carissa
  Schoenick, and Oyvind Tafjord.
\newblock Think you have solved question answering? try arc, the ai2 reasoning
  challenge, 2018.
\newblock URL \url{https://arxiv.org/abs/1803.05457}.

\bibitem[De et~al.(2024)De, Smith, Fernando, Botev, Cristian-Muraru, Gu,
  Haroun, Berrada, Chen, Srinivasan, Desjardins, Doucet, Budden, Teh, Pascanu,
  Freitas, and Gulcehre]{de2024griffinmixinggatedlinear}
Soham De, Samuel~L. Smith, Anushan Fernando, Aleksandar Botev, George
  Cristian-Muraru, Albert Gu, Ruba Haroun, Leonard Berrada, Yutian Chen,
  Srivatsan Srinivasan, Guillaume Desjardins, Arnaud Doucet, David Budden,
  Yee~Whye Teh, Razvan Pascanu, Nando~De Freitas, and Caglar Gulcehre.
\newblock Griffin: Mixing gated linear recurrences with local attention for
  efficient language models, 2024.
\newblock URL \url{https://arxiv.org/abs/2402.19427}.

\bibitem[DeepSeek-AI(2025)]{deepseekai2025deepseekr1incentivizingreasoningcapability}
DeepSeek-AI.
\newblock Deepseek-r1: Incentivizing reasoning capability in llms via
  reinforcement learning, 2025.
\newblock URL \url{https://arxiv.org/abs/2501.12948}.

\bibitem[Dong et~al.(2024)Dong, Fu, Diao, Byeon, Chen, Mahabaleshwarkar, Liu,
  Keirsbilck, Chen, Suhara, Lin, Kautz, and
  Molchanov]{dong2024hymbahybridheadarchitecturesmall}
Xin Dong, Yonggan Fu, Shizhe Diao, Wonmin Byeon, Zijia Chen, Ameya~Sunil
  Mahabaleshwarkar, Shih-Yang Liu, Matthijs~Van Keirsbilck, Min-Hung Chen,
  Yoshi Suhara, Yingyan Lin, Jan Kautz, and Pavlo Molchanov.
\newblock Hymba: A hybrid-head architecture for small language models, 2024.
\newblock URL \url{https://arxiv.org/abs/2411.13676}.

\bibitem[Elfwing et~al.(2017)Elfwing, Uchibe, and
  Doya]{elfwing2017sigmoidweightedlinearunitsneural}
Stefan Elfwing, Eiji Uchibe, and Kenji Doya.
\newblock Sigmoid-weighted linear units for neural network function
  approximation in reinforcement learning, 2017.
\newblock URL \url{https://arxiv.org/abs/1702.03118}.

\bibitem[Fang et~al.(2024)Fang, Wang, Liu, Zhang, Jegelka, Gao, Ding, and
  Wang]{fang2025wrongperplexitylongcontextlanguage}
Lizhe Fang, Yifei Wang, Zhaoyang Liu, Chenheng Zhang, Stefanie Jegelka, Jinyang
  Gao, Bolin Ding, and Yisen Wang.
\newblock What is wrong with perplexity for long-context language modeling?,
  2024.
\newblock URL \url{https://arxiv.org/abs/2410.23771}.

\bibitem[Fu et~al.(2023)Fu, Dao, Saab, Thomas, Rudra, and
  Ré]{fu2023hungryhungryhipposlanguage}
Daniel~Y. Fu, Tri Dao, Khaled~K. Saab, Armin~W. Thomas, Atri Rudra, and
  Christopher Ré.
\newblock Hungry hungry hippos: Towards language modeling with state space
  models, 2023.
\newblock URL \url{https://arxiv.org/abs/2212.14052}.

\bibitem[Gehring et~al.(2025)Gehring, Zheng, Copet, Mella, Carbonneaux, Cohen,
  and Synnaeve]{gehring2025rlefgroundingcodellms}
Jonas Gehring, Kunhao Zheng, Jade Copet, Vegard Mella, Quentin Carbonneaux,
  Taco Cohen, and Gabriel Synnaeve.
\newblock Rlef: Grounding code llms in execution feedback with reinforcement
  learning, 2025.
\newblock URL \url{https://arxiv.org/abs/2410.02089}.

\bibitem[Gemma~Team(2025)]{gemmateam2025gemma3technicalreport}
Google~DeepMind Gemma~Team.
\newblock Gemma 3 technical report, 2025.
\newblock URL \url{https://arxiv.org/abs/2503.19786}.

\bibitem[Gu \& Dao(2024)Gu and Dao]{gu2024mambalineartimesequencemodeling}
Albert Gu and Tri Dao.
\newblock Mamba: Linear-time sequence modeling with selective state spaces,
  2024.
\newblock URL \url{https://arxiv.org/abs/2312.00752}.

\bibitem[Hochreiter \& Schmidhuber(1997)Hochreiter and Schmidhuber]{LSTM}
Sepp Hochreiter and Jürgen Schmidhuber.
\newblock Long short-term memory.
\newblock \emph{Neural Computation}, 9:\penalty0 1735--1780, 11 1997.
\newblock \doi{10.1162/neco.1997.9.8.1735}.

\bibitem[Hsieh et~al.(2024)Hsieh, Sun, Kriman, Acharya, Rekesh, Jia, Zhang, and
  Ginsburg]{hsieh2024rulerwhatsrealcontext}
Cheng-Ping Hsieh, Simeng Sun, Samuel Kriman, Shantanu Acharya, Dima Rekesh, Fei
  Jia, Yang Zhang, and Boris Ginsburg.
\newblock Ruler: What's the real context size of your long-context language
  models?, 2024.
\newblock URL \url{https://arxiv.org/abs/2404.06654}.

\bibitem[Jain et~al.(2024)Jain, Han, Gu, Li, Yan, Zhang, Wang, Solar-Lezama,
  Sen, and Stoica]{jain2024livecodebench}
Naman Jain, King Han, Alex Gu, Wen-Ding Li, Fanjia Yan, Tianjun Zhang, Sida
  Wang, Armando Solar-Lezama, Koushik Sen, and Ion Stoica.
\newblock Livecodebench: Holistic and contamination free evaluation of large
  language models for code.
\newblock \emph{arXiv preprint}, 2024.

\bibitem[Joshi et~al.(2017)Joshi, Choi, Weld, and
  Zettlemoyer]{joshi-etal-2017-triviaqa}
Mandar Joshi, Eunsol Choi, Daniel Weld, and Luke Zettlemoyer.
\newblock {T}rivia{QA}: A large scale distantly supervised challenge dataset
  for reading comprehension.
\newblock In Regina Barzilay and Min-Yen Kan (eds.), \emph{Proceedings of the
  55th Annual Meeting of the Association for Computational Linguistics (Volume
  1: Long Papers)}, pp.\  1601--1611, Vancouver, Canada, July 2017. Association
  for Computational Linguistics.
\newblock \doi{10.18653/v1/P17-1147}.
\newblock URL \url{https://aclanthology.org/P17-1147/}.

\bibitem[Katharopoulos et~al.(2020)Katharopoulos, Vyas, Pappas, and
  Fleuret]{katharopoulos2020transformersrnnsfastautoregressive}
Angelos Katharopoulos, Apoorv Vyas, Nikolaos Pappas, and François Fleuret.
\newblock Transformers are rnns: Fast autoregressive transformers with linear
  attention, 2020.
\newblock URL \url{https://arxiv.org/abs/2006.16236}.

\bibitem[Kuratov et~al.(2024)Kuratov, Bulatov, Anokhin, Rodkin, Sorokin,
  Sorokin, and Burtsev]{kuratov2024babilong}
Yuri Kuratov, Aydar Bulatov, Petr Anokhin, Ivan Rodkin, Dmitry~Igorevich
  Sorokin, Artyom Sorokin, and Mikhail Burtsev.
\newblock {BABIL}ong: Testing the limits of {LLM}s with long context
  reasoning-in-a-haystack.
\newblock In \emph{The Thirty-eight Conference on Neural Information Processing
  Systems Datasets and Benchmarks Track}, 2024.
\newblock URL \url{https://openreview.net/forum?id=u7m2CG84BQ}.

\bibitem[Kwiatkowski et~al.(2019)Kwiatkowski, Palomaki, Redfield, Collins,
  Parikh, Alberti, Epstein, Polosukhin, Devlin, Lee, Toutanova, Jones, Kelcey,
  Chang, Dai, Uszkoreit, Le, and Petrov]{kwiatkowski-etal-2019-natural}
Tom Kwiatkowski, Jennimaria Palomaki, Olivia Redfield, Michael Collins, Ankur
  Parikh, Chris Alberti, Danielle Epstein, Illia Polosukhin, Jacob Devlin,
  Kenton Lee, Kristina Toutanova, Llion Jones, Matthew Kelcey, Ming-Wei Chang,
  Andrew~M. Dai, Jakob Uszkoreit, Quoc Le, and Slav Petrov.
\newblock Natural questions: A benchmark for question answering research.
\newblock \emph{Transactions of the Association for Computational Linguistics},
  7:\penalty0 452--466, 2019.
\newblock \doi{10.1162/tacl_a_00276}.
\newblock URL \url{https://aclanthology.org/Q19-1026/}.

\bibitem[Lai et~al.(2017)Lai, Xie, Liu, Yang, and
  Hovy]{lai2017racelargescalereadingcomprehension}
Guokun Lai, Qizhe Xie, Hanxiao Liu, Yiming Yang, and Eduard Hovy.
\newblock Race: Large-scale reading comprehension dataset from examinations,
  2017.
\newblock URL \url{https://arxiv.org/abs/1704.04683}.

\bibitem[Liu et~al.(2024)Liu, Wang, Wu, Feng, Stone, and
  Liu]{liu2024longhornstatespacemodels}
Bo~Liu, Rui Wang, Lemeng Wu, Yihao Feng, Peter Stone, and Qiang Liu.
\newblock Longhorn: State space models are amortized online learners, 2024.
\newblock URL \url{https://arxiv.org/abs/2407.14207}.

\bibitem[Liu et~al.(2021)Liu, Dai, So, and Le]{liu2021payattentionmlps}
Hanxiao Liu, Zihang Dai, David~R. So, and Quoc~V. Le.
\newblock Pay attention to mlps, 2021.
\newblock URL \url{https://arxiv.org/abs/2105.08050}.

\bibitem[Liu et~al.(2023)Liu, Xia, Wang, and Zhang]{evalplus}
Jiawei Liu, Chunqiu~Steven Xia, Yuyao Wang, and Lingming Zhang.
\newblock Is your code generated by chat{GPT} really correct? rigorous
  evaluation of large language models for code generation.
\newblock In \emph{Thirty-seventh Conference on Neural Information Processing
  Systems}, 2023.
\newblock URL \url{https://openreview.net/forum?id=1qvx610Cu7}.

\bibitem[NVIDIA(2025)]{nvidia2025nemotronhfamilyaccurateefficient}
Team NVIDIA.
\newblock Nemotron-h: A family of accurate and efficient hybrid
  mamba-transformer models, 2025.
\newblock URL \url{https://arxiv.org/abs/2504.03624}.

\bibitem[OpenAI(2025)]{openai2025gptoss120bgptoss20bmodel}
OpenAI.
\newblock gpt-oss-120b \& gpt-oss-20b model card, 2025.
\newblock URL \url{https://arxiv.org/abs/2508.10925}.

\bibitem[Peng et~al.(2023)Peng, Quesnelle, Fan, and
  Shippole]{peng2023yarnefficientcontextwindow}
Bowen Peng, Jeffrey Quesnelle, Honglu Fan, and Enrico Shippole.
\newblock Yarn: Efficient context window extension of large language models,
  2023.
\newblock URL \url{https://arxiv.org/abs/2309.00071}.

\bibitem[Ren et~al.(2025)Ren, Liu, Lu, Shen, Liang, and
  Chen]{ren2025sambasimplehybridstate}
Liliang Ren, Yang Liu, Yadong Lu, Yelong Shen, Chen Liang, and Weizhu Chen.
\newblock Samba: Simple hybrid state space models for efficient unlimited
  context language modeling, 2025.
\newblock URL \url{https://arxiv.org/abs/2406.07522}.

\bibitem[Ruiz \& Gu(2025)Ruiz and
  Gu]{ruiz2025understandingimprovinglengthgeneralization}
Ricardo~Buitrago Ruiz and Albert Gu.
\newblock Understanding and improving length generalization in recurrent
  models, 2025.
\newblock URL \url{https://arxiv.org/abs/2507.02782}.

\bibitem[Sakaguchi et~al.(2019)Sakaguchi, Bras, Bhagavatula, and
  Choi]{sakaguchi2019winograndeadversarialwinogradschema}
Keisuke Sakaguchi, Ronan~Le Bras, Chandra Bhagavatula, and Yejin Choi.
\newblock Winogrande: An adversarial winograd schema challenge at scale, 2019.
\newblock URL \url{https://arxiv.org/abs/1907.10641}.

\bibitem[Sap et~al.(2019)Sap, Rashkin, Chen, LeBras, and
  Choi]{sap2019socialiqacommonsensereasoningsocial}
Maarten Sap, Hannah Rashkin, Derek Chen, Ronan LeBras, and Yejin Choi.
\newblock Socialiqa: Commonsense reasoning about social interactions, 2019.
\newblock URL \url{https://arxiv.org/abs/1904.09728}.

\bibitem[Singh et~al.(2026)Singh, Krauss, Jaghouar, Sirovatka, Goddard, Obied,
  Ong, Straube, Fern, Harley, Stewart, Kealty, Panahi, Kirsten, Deshpande, Vij,
  Bresnu, Veldurthi, Ravishankar, Bishnoi, Team, Team, Team, McQuade, Hagemann,
  and Atkins]{singh2026arceetrinity}
Varun Singh, Lucas Krauss, Sami Jaghouar, Matej Sirovatka, Charles Goddard,
  Fares Obied, Jack~Min Ong, Jannik Straube, Fern, Aria Harley, Conner Stewart,
  Colin Kealty, Maziyar Panahi, Simon Kirsten, Anushka Deshpande, Anneketh Vij,
  Arthur Bresnu, Pranav Veldurthi, Raghav Ravishankar, Hardik Bishnoi,
  DatologyAI Team, Arcee~AI Team, Prime~Intellect Team, Mark McQuade, Johannes
  Hagemann, and Lucas Atkins.
\newblock Arcee trinity large technical report, 2026.
\newblock URL \url{https://arxiv.org/abs/2602.17004}.

\bibitem[Srivastava et~al.(2014)Srivastava, Hinton, Krizhevsky, Sutskever, and
  Salakhutdinov]{JMLR:v15:srivastava14a}
Nitish Srivastava, Geoffrey Hinton, Alex Krizhevsky, Ilya Sutskever, and Ruslan
  Salakhutdinov.
\newblock Dropout: A simple way to prevent neural networks from overfitting.
\newblock \emph{Journal of Machine Learning Research}, 15\penalty0
  (56):\penalty0 1929--1958, 2014.
\newblock URL \url{http://jmlr.org/papers/v15/srivastava14a.html}.

\bibitem[Su et~al.(2023)Su, Lu, Pan, Murtadha, Wen, and
  Liu]{su2023roformerenhancedtransformerrotary}
Jianlin Su, Yu~Lu, Shengfeng Pan, Ahmed Murtadha, Bo~Wen, and Yunfeng Liu.
\newblock Roformer: Enhanced transformer with rotary position embedding, 2023.
\newblock URL \url{https://arxiv.org/abs/2104.09864}.

\bibitem[Sun et~al.(2025)Sun, Li, Dalal, Xu, Vikram, Zhang, Dubois, Chen, Wang,
  Koyejo, Hashimoto, and Guestrin]{sun2025learninglearntesttime}
Yu~Sun, Xinhao Li, Karan Dalal, Jiarui Xu, Arjun Vikram, Genghan Zhang, Yann
  Dubois, Xinlei Chen, Xiaolong Wang, Sanmi Koyejo, Tatsunori Hashimoto, and
  Carlos Guestrin.
\newblock Learning to (learn at test time): Rnns with expressive hidden states,
  2025.
\newblock URL \url{https://arxiv.org/abs/2407.04620}.

\bibitem[Wang et~al.(2025)Wang, Zhu, Abreu, Shan, Kergan, Pan, Chou, Li, Zhang,
  Huang, and Eshraghian]{wang2025systematicanalysishybridlinear}
Dustin Wang, Rui-Jie Zhu, Steven Abreu, Yong Shan, Taylor Kergan, Yuqi Pan,
  Yuhong Chou, Zheng Li, Ge~Zhang, Wenhao Huang, and Jason Eshraghian.
\newblock A systematic analysis of hybrid linear attention, 2025.
\newblock URL \url{https://arxiv.org/abs/2507.06457}.

\bibitem[Wei et~al.(2023)Wei, Wang, Schuurmans, Bosma, Ichter, Xia, Chi, Le,
  and Zhou]{wei2023chainofthoughtpromptingelicitsreasoning}
Jason Wei, Xuezhi Wang, Dale Schuurmans, Maarten Bosma, Brian Ichter, Fei Xia,
  Ed~Chi, Quoc Le, and Denny Zhou.
\newblock Chain-of-thought prompting elicits reasoning in large language
  models, 2023.
\newblock URL \url{https://arxiv.org/abs/2201.11903}.

\bibitem[Weston et~al.(2015)Weston, Bordes, Chopra, and
  Mikolov]{Weston2015TowardsAQ}
Jason Weston, Antoine Bordes, Sumit Chopra, and Tomas Mikolov.
\newblock Towards ai-complete question answering: A set of prerequisite toy
  tasks.
\newblock \emph{arXiv: Artificial Intelligence}, 2015.
\newblock URL \url{https://api.semanticscholar.org/CorpusID:3178759}.

\bibitem[Xiao(2025)]{xiao2025sliding}
Guangxuan Xiao.
\newblock Why stacking sliding windows can't see very far.
\newblock \url{https://guangxuanx.com/blog/stacking-swa.html}, 2025.

\bibitem[Yang et~al.(2024)Yang, Wang, Shen, Panda, and
  Kim]{yang2024gatedlinearattentiontransformers}
Songlin Yang, Bailin Wang, Yikang Shen, Rameswar Panda, and Yoon Kim.
\newblock Gated linear attention transformers with hardware-efficient training,
  2024.
\newblock URL \url{https://arxiv.org/abs/2312.06635}.

\bibitem[Yang et~al.(2025)Yang, Kautz, and
  Hatamizadeh]{yang2025gateddeltanetworksimproving}
Songlin Yang, Jan Kautz, and Ali Hatamizadeh.
\newblock Gated delta networks: Improving mamba2 with delta rule, 2025.
\newblock URL \url{https://arxiv.org/abs/2412.06464}.

\bibitem[Zellers et~al.(2019)Zellers, Holtzman, Bisk, Farhadi, and
  Choi]{zellers2019hellaswagmachinereallyfinish}
Rowan Zellers, Ari Holtzman, Yonatan Bisk, Ali Farhadi, and Yejin Choi.
\newblock Hellaswag: Can a machine really finish your sentence?, 2019.
\newblock URL \url{https://arxiv.org/abs/1905.07830}.

\bibitem[Zhang \& Sennrich(2019)Zhang and
  Sennrich]{zhang2019rootmeansquarelayer}
Biao Zhang and Rico Sennrich.
\newblock Root mean square layer normalization, 2019.
\newblock URL \url{https://arxiv.org/abs/1910.07467}.

\bibitem[Zhang \& Bottou(2025)Zhang and Bottou]{zhang2025memory}
Jianyu Zhang and L{\'e}on Bottou.
\newblock Memory mosaics at scale.
\newblock \emph{arXiv preprint arXiv:2507.03285}, 2025.

\bibitem[Zhong et~al.(2025)Zhong, Xu, Ao, and
  Shi]{zhong2025understandingtransformerperspectiveassociative}
Shu Zhong, Mingyu Xu, Tenglong Ao, and Guang Shi.
\newblock Understanding transformer from the perspective of associative memory,
  2025.
\newblock URL \url{https://arxiv.org/abs/2505.19488}.

\end{thebibliography}
